\documentclass[11pt]{article}

\usepackage[preprint]{acl}
\usepackage{booktabs}
\usepackage{multirow}
\usepackage{enumitem}
\usepackage{tcolorbox}
\usepackage{amsmath}
\usepackage{amssymb}
\usepackage{times}
\usepackage{latexsym}
\usepackage{amsmath}
\usepackage{caption}
\usepackage{subcaption}
\usepackage[T1]{fontenc}

\usepackage[utf8]{inputenc}

\usepackage{microtype}

\usepackage{inconsolata}
\usepackage{graphicx}

\usepackage{algorithm}
\usepackage{algorithmic}

\usepackage{ifthen}

\newboolean{showcomments}
\setboolean{showcomments}{false} 

\newcommand{\sh}[1]{\ifthenelse{\boolean{showcomments}}{\textcolor{blue}{[Songyang: #1]}}{}}
\newcommand{\joe}[1]{\ifthenelse{\boolean{showcomments}}{\textcolor{red}{[JH: #1]}}{}}
\newcommand{\ms}[1]{\ifthenelse{\boolean{showcomments}}{\textcolor{magenta}{[Manan: #1]}}{}}
\newcommand{\xiangci}[1]{\ifthenelse{\boolean{showcomments}}{\textcolor{green}{[XL: #1]}}{}}

\usepackage[page,header]{appendix}
\usepackage{titletoc}

\title{CodeScout: Contextual Problem Statement Enhancement \\ for Software Agents}

\author{
Manan Suri$^{1\thanks{Work done while at Amazon.}}$, 
Xiangci Li$^{2\thanks{Corresponding author.}}$, 
Mehdi Shojaie$^{2}$, 
Songyang Han$^{2}$, 
Chao-Chun Hsu$^{2}$, 
\\
\textbf{Shweta Garg}$^{2}$, 
\textbf{Aniket Deshmukh}$^{3\footnotemark[1]}$, 
\textbf{Varun Kumar}$^{2}$ \\
\\
$^{1}$University of Maryland, College Park \hspace{10pt}
$^{2}$Amazon Web Services (AWS) \hspace{10pt}
$^{3}$Databricks \\
\\
\texttt{manans@umd.edu}, \texttt{xiangcil@amazon.com}
}

\begin{document}
\maketitle
\begin{abstract}
Current AI-powered code assistance tools often struggle with poorly-defined problem statements that lack sufficient task context and requirements specification. Recent analysis of software engineering agents reveals that failures on such underspecified requests are highly correlated with longer trajectories involving either over-exploration or repeated attempts at applying the same fix without proper evolution or testing, leading to suboptimal outcomes across software development tasks. We introduce CodeScout, a contextual query refinement approach that systematically converts underspecified user requests into comprehensive, actionable problem statements through lightweight pre-exploration of the target codebase. Our key innovation is demonstrating that structured analysis before task execution can supplement existing agentic capabilities without requiring any modifications to their underlying scaffolds. CodeScout performs targeted context scoping, conducts multi-perspective analysis examining potential fixes and exploration opportunities, then synthesizes these insights into enhanced problem statements with reproduction steps, expected behaviors, and targeted exploration hints. This pre-exploration directly addresses the identified failure patterns by reducing non-converging agent trajectories while clarifying user intent in natural language space.
We evaluate CodeScout using state-of-the-art agentic scaffolds and language models on SWEBench-Verified, demonstrating a 20\% improvement in resolution rates with up to 27 additional issues resolved compared to the default baseline method. Our results suggest that systematic query refinement through contextual analysis represents a promising direction for enhancing AI code assistance capabilities.
\end{abstract}

\section{Introduction}
\begin{figure*}
    \centering
    \includegraphics[width=\linewidth]{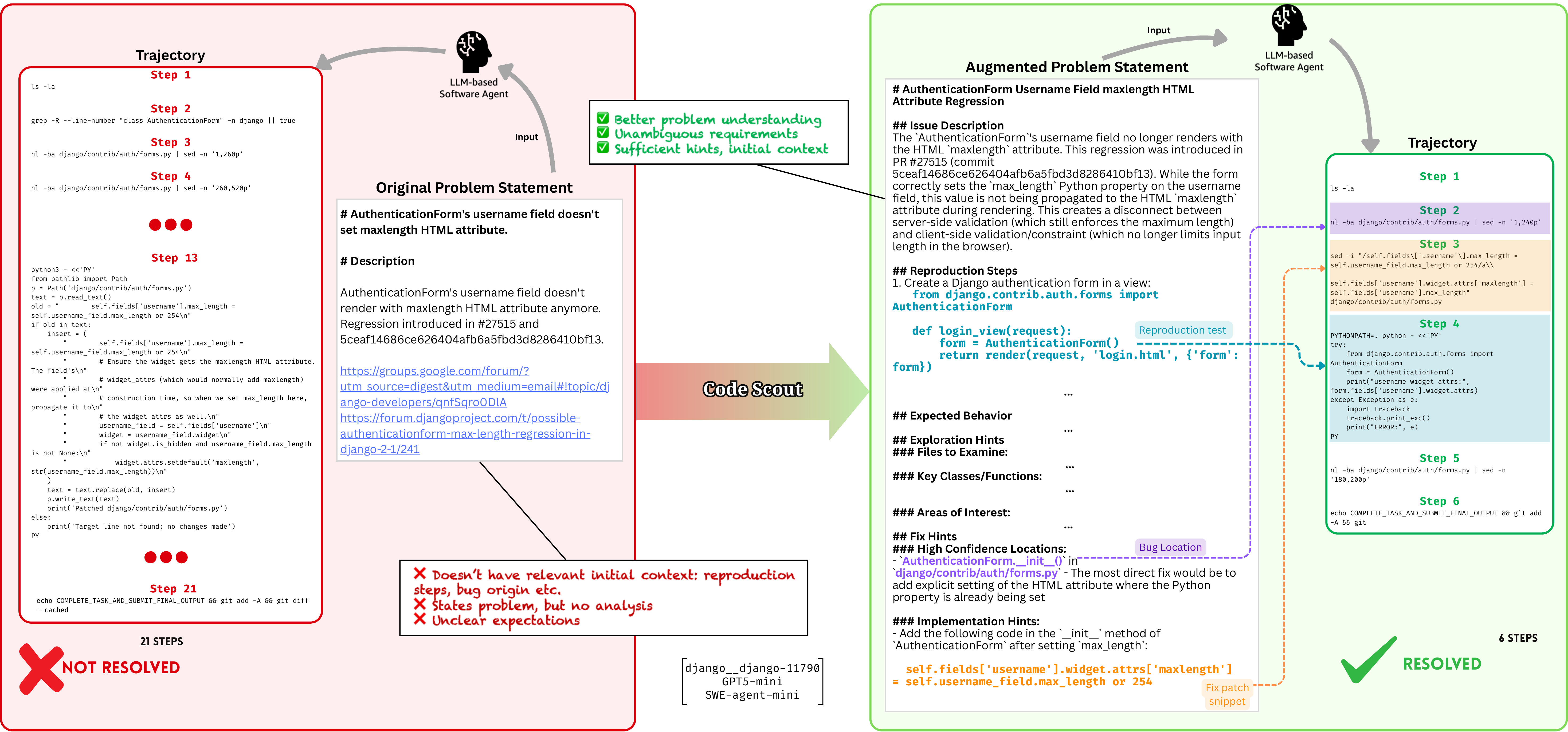}
    \caption{\small{The original SWEBench problem statement for Instance \texttt{django\_\_django-11790} lack relevant initial context. As a result, the downstream agent is not able to fix the issue, despite spending 21 steps exploring the repository, analyzing code, iterating the fix patch. In contrast, the enhanced problem statement generated by our approach resolves the issue in 6 agentic steps, as it includes relevant insights that can be used off-the-bat by the agent.}}
    \label{fig:motivation}
\end{figure*}

The rapid advancement of large language models (LLMs) has revolutionized software development assistance, with AI-powered coding tools becoming increasingly integral to modern workflows \cite{kumar2025intuitionevidencemeasuringais, jiang2023impact}. Yet a fundamental challenge persists: these systems often fail not because the underlying models lack reasoning capability, but because they are asked to operate on poorly specified problem statements \cite{meng2024empirical}. Developers frequently provide concise, context-dependent descriptions—omitting reproduction steps, technical details, or clear expectations—that assume a shared understanding of the codebase. In contrast, LLM-based agents work optimally with explicit, well-scoped specifications to reason effectively \cite{nam2025prompting}. \sh{If we have experiment results, consider to add a teaser figure here to show the improvement if LLM can work with well-scoped specifications.} \ms{Figure~\ref{fig:motivation} discussion + mentioned stats in contributions}

Empirical studies validate this phenomenon: \citet{meng2024empirical} found that resolvable bug reports exhibit drastically higher description quality scores, with relative differences ranging from 110\% to over 2700\% compared to non-resolvable ones. Similarly, \citet{nam2025prompting} showed that rejected code suggestions in Google's LLM-integrated development tools strongly correlate with low-quality user inputs. Recent analysis by \citet{bouzenia2025understanding} reveals that software engineering agent failures exhibit specific behavioral patterns: over-exploration where agents fail to reach the root of the problem due to context overload, and repeated application of the same fix without proper testing or evolution, demonstrating stubborn agentic behavior and lack of understanding of code or user intent. Together, these findings highlight that input quality—not just model capacity—forms the critical bottleneck in AI-assisted software engineering.

This specification gap is exacerbated by how current agents operate, but why do current agentic workflows struggle with this challenge? We hypothesize this occurs for several fundamental reasons. First, agents plan and implement simultaneously at a step-by-step granularity through reason-execute-observe loops, lacking the long-horizon understanding necessary for complex debugging tasks and accumulating deviations from the true problem scope. Second, without comprehensive upfront context, agents become trapped in reactive exploration, facing information asymmetry where they must discover codebase structure incrementally while lacking the hierarchical abstraction capabilities to build strategic mental models. Figure~\ref{fig:motivation} \xiangci{The words in this figure are too small to read} \ms{Will increase font size in figure revision.} exemplifies this: a sparse Django bug report leads to 21 steps of unfocused exploration and failure, while an enhanced problem statement enables success in just 6 targeted steps. \sh{We need to answer some questions in intro: 1. Why agentic workflow cannot solve it? Why we need to enforce the problem understanding as a plug-in module. 2. What is the difference with localization or context retrieval.} \ms{Addressed: Agentic workflows struggle due to short-term reasoning loops and information asymmetry. Unlike localization/retrieval, we provide plans, task decomposition, and strategic guidance.}

This observation motivates a shift in perspective: instead of expecting agents to "leap" directly into problem-solving, we argue they must first look—building comprehensive understanding before attempting fixes. We introduce CodeScout, a contextual problem statement enhancement method, that transforms vague developer queries into rich, repository-aware actionable problem statements. Compared to traditional localization or context retrieval methods that merely identify relevant code, CodeScout provides detailed exploration plans, task decomposition, fix hints, and strategic guidance. Key advantages include: supplementing existing agents without requiring scaffold modifications, enabling long-horizon strategic alignment, reducing non-converging trajectories, and working universally across agent architectures. \textbf{Main contributions}:


\begin{enumerate}
\item CodeScout, a systematic approach for contextual problem specification enhancement that improves input quality through repository-aware analysis, demonstrating 20\% improvement in resolution rates with up to 27 additional issues resolved.
\item Empirical validation demonstrating CodeScout's effectiveness across multiple software engineering tasks and agent architectures, showing consistent performance gains.
\item Detailed analysis of how CodeScout's contextual augmentation influences agent trajectories, tool usage, and problem-solving strategies, revealing cross-validation performance gains that generalize across different agent implementations. We demonstrate compelling cost-efficiency dynamics: smaller models can effectively perform problem statement enhancement to improve larger model performance, while strong models can augment problem statements to dramatically boost weaker model capabilities. \sh{Highlight some important and attractive take-aways here. For example, the results from cross-validation.} \ms{Updated.}
\end{enumerate}

Our results suggest that investing computation in upfront problem understanding offers a powerful complement to advances in model capacity and agentic reasoning, pointing toward more reliable AI-assisted software engineering.

\section{Related Work}
\sh{You miss to introduce how these work related to our work. Do they help shape the idea of this paper? What is the difference with our work? Is there any challenge they cannot solve but solved by our design?} \ms{Updated}

\textbf{LLM-based Software Agents}

LLM-based agents \cite{wang2024srsa, romeo2025arpaccino} have emerged as autonomous systems that use large language models as reasoning engines to perform complex software engineering tasks through planning, tool usage, and iterative problem-solving. These agents substantially extend the versatility of LLMs by enhancing them with capabilities for perceiving and utilizing external resources and tools, providing autonomy through key components including planning, memory, perception, and action \cite{liu2024large}. The sophisticated capabilities demonstrated by these agents directly motivated our work: while they excel at execution and reasoning, their performance degrades significantly when operating on poorly specified problem statements—a fundamental limitation that CodeScout addresses.

The evolution from traditional LLM applications to autonomous agents represents a significant shift in software engineering automation. While earlier approaches typically operated through single prompts or hard-coded feedback loops for tasks like code completion \cite{ziegler2022productivity, barke2023grounded}, automated program repair \cite{jiang2023impact}, and test generation \cite{lemieux2023codamosa, ryan2024code, yuan2024evaluating}, modern LLM agents \cite{bouzenia2025understanding} can autonomously plan and execute sequences of actions while adapting based on feedback from tools and environments.

Several prominent agent frameworks have demonstrated effectiveness across software engineering tasks, but also revealed the input quality bottleneck that CodeScout targets. RepairAgent \cite{bouzenia2025repairagent} treats the LLM as an autonomous agent capable of planning and executing actions to fix bugs by invoking suitable tools, freely interleaving information gathering, repair ingredient collection, and fix validation. AutoCodeRover \cite{zhang2024autocoderover} combines LLMs with sophisticated code search capabilities, working on program representations like abstract syntax trees rather than viewing projects as mere file collections. SWE-agent \cite{yang2024sweagent} introduces custom agent-computer interfaces that significantly enhance agents' abilities to create, edit, navigate repositories, and execute tests. \xiangci{nit: put citation right after the prior work name.} \ms{Fixed citation placement throughout.}

However, these agents share a critical limitation: they cannot compensate for fundamentally under-specified input problems. Unlike CodeScout, which proactively enhances problem understanding before agent execution, existing agents rely on reactive exploration that often leads to the over-exploration and repeated fix application patterns identified by recent analysis \cite{bouzenia2025understanding}.

\begin{figure*}[h]
    \centering
    \includegraphics[width=\linewidth]{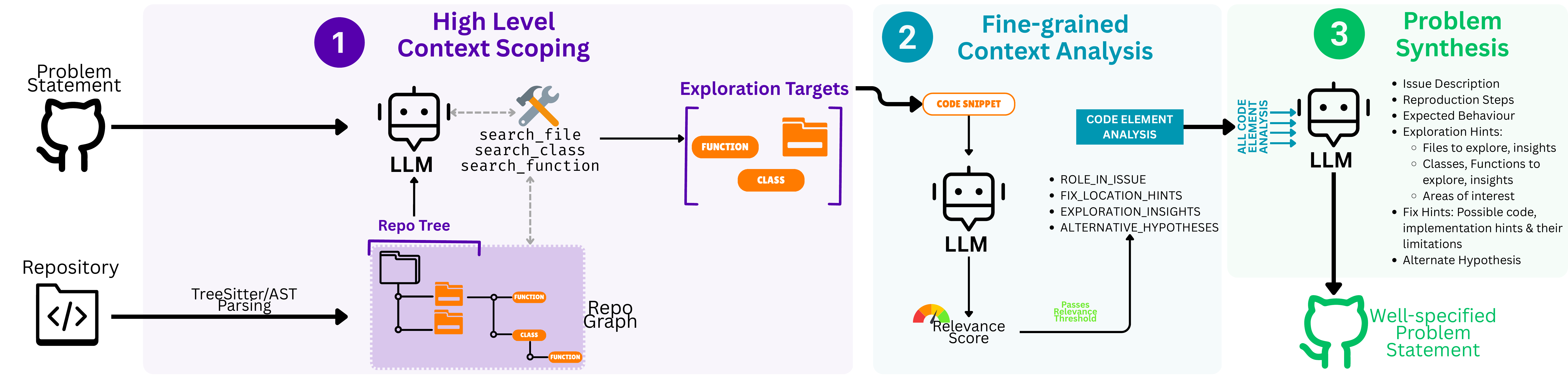}
    \caption{\small{\textbf{CodeScout:} The pre-exploration with Repository Knowledge Graph Construction, which represents code structure and relationships. Building on this, three main stages follow: 1) \textit{High Level Scoping}, where an LLM agent identifies relevant exploration targets, 2) \textit{Fine-grained Context Analysis}, which extracts structured insights for each target, and 3)\textit{ Problem Synthesis}, where the original problem statement is combined with filtered insights to generate the augmented specification.}}
    \label{fig:main}
\end{figure*}

\textbf{Code Query Understanding and Refinement}

While LLM-based software agents demonstrate sophisticated capabilities for repository-level interactions, existing approaches to code query enhancement have primarily focused on traditional information retrieval techniques without addressing the semantic understanding of complex problem statements at the repository level. This gap in repository-aware problem understanding directly shaped the development of CodeScout's contextual enhancement approach.

Traditional query expansion approaches have relied on lexical methods to address the vocabulary mismatch problem between queries and code. Lu et al. \cite{lu2015query} proposed extending queries with synonyms generated from WordNet to match natural language phrases extracted from source code identifiers. Nie et al. \cite{nie2016query} introduced Query Expansion based on Crowd Knowledge (QECK), which identifies software-specific expansion words from Stack Overflow question-answer pairs to automatically generate expansion queries, improving precision by up to 64\%. While these approaches improve keyword matching, they cannot handle the complex, repository-contextual problem statements that CodeScout addresses.

Recent work has explored LLM-based data augmentation for code search. Wang et al. \cite{wang2023you} proposed ChatDANCE, which utilizes ChatGPT to generate high-quality augmented code-query pairs with a filtering mechanism, achieving 13.2\% improvement in R@1. However, this approach focuses on augmenting training data for code search rather than understanding complex problem specifications in live development contexts.

In program repair contexts, some work has explored semantic augmentation of prompts. Ahmed et al. \cite{ahmed2024automatic} demonstrated that adding semantic facts (parameter names, control flow, etc.) to LLM prompts improves code summarization performance, surpassing 30 BLEU1 on PHP in CodeSearchNet. Jin et al. \cite{jin2023inferfixendtoendprogramrepair} proposed InferFix, which combines retrieval of semantically similar bug-fix pairs with LLM-based generation, achieving 76.8\% accuracy in Java program repair. These approaches inform our semantic analysis component but lack the comprehensive repository-aware problem statement synthesis that CodeScout provides. The fundamental challenge that existing approaches cannot solve—and that motivated CodeScout's design—is transforming vague, context-dependent developer descriptions into comprehensive, actionable problem statements that enable strategic rather than reactive agent behavior through repository-level understanding with cross-file dependencies, architectural context, and hierarchical refinement that mirrors human developer comprehension.

\section{Task Definition}

We formalize the problem statement augmentation task as a preprocessing transformation that enhances task specifications for downstream software agents. Given an initial problem statement $P_0$, software repository $\mathcal{R}$, and task-specific agent $\mathcal{A}$, the repository contains implicit contextual information. Through systematic codebase analysis, we extract structured contextual knowledge $\mathcal{S}$ from $\mathcal{R}$. Our objective is to learn a transformation function $\mathcal{T}$ such that:

\begin{equation}
   M(\mathcal{A}(\mathcal{P}_{aug}, \mathcal{R})) \geq M(\mathcal{A}(P_0, \mathcal{R})) 
\end{equation}

where $P_{aug} = \mathcal{T}(P_0, \mathcal{R}, \mathcal{S})$, represents the augmented specification that enriches the original problem statement with extracted contextual knowledge, $M(\cdot)$ is the task performance metric. The transformation $\mathcal{T}$ performs context retrieval and structuring from $\mathcal{R}$, making implicit information explicit to reduce the agent's computational burden. This approach operates as a plug-and-play preprocessing step, decoupled from downstream agent architectures while providing enhanced task specifications.

\section{CodeScout}
We now describe the implementation of $\mathcal{T}(P_0, \mathcal{R}, \mathcal{S})$ through \textbf{CodeScout} (Figure~\ref{fig:main}), our three-stage pipeline that systematically extracts structured knowledge $\mathcal{S}$ from the repository and synthesizes augmented problem specifications. The pipeline operates as follows: (1) Repository analysis constructs a knowledge graph $G(\mathcal{R})$ representing code structure and relationships, (2) Context scoping identifies relevant exploration targets $T$ from $G(\mathcal{R})$ and $P_0$, (3) Code analysis extracts technical insights $I$ for each target, and (4) Specification synthesis combines $P_0$ with the extracted knowledge to produce $P_{aug}$. Together, the knowledge graph $G(\mathcal{R})$, exploration targets $T$, and insights $I$ constitute the structured contextual knowledge $\mathcal{S}$ from our task formulation (Section 3).

\subsection{Repository Knowledge Graph Construction}

We construct a directed graph $G(\mathcal{R}) = (V, E)$ where vertices $V$ represent code entities and edges $E$ capture semantic relationships. Each vertex $v \in V$ is characterized by its name, type, location, and metadata. We employ an Abstract Syntax Tree (AST) visitor that traverses parsed syntax trees to extract class definitions with inheritance relationships, function signatures with parameter specifications, import dependencies and module relationships, and variable declarations and scope information.

\subsection{High Level Scoping}

Given problem statement $P_0$ and the repository graph $G(\mathcal{R})$, an LLM agent generates exploration targets:

\begin{equation}
    T = \text{LLM}_{scope}(P_0, G(\mathcal{R}))
\end{equation}

The scoping agent identifies code entities most likely relevant to the reported issue by analyzing direct entity mentions in $P_0$, applying common debugging patterns from its training knowledge, examining structural relationships within $G(\mathcal{R})$, and considering entity naming conventions that might relate to the issue description.

The repository structure representation $G(\mathcal{R})$ presents hierarchical organization with semantic annotations showing both file organization and code structure, including the complete file tree layout. During this scoping phase, the actual retrieval of relevant entities occurs directly from this repository-graph, enabling efficient identification of potential targets without requiring complete source code access.

Exploration targets are constrained to $|T| \leq 15$ to balance comprehensive coverage with computational feasibility. Each target $t_i$ includes the entity type, name, and reasoning explaining its relevance to $P_0$.

\subsection{Fine-grained Context Analysis}

For each exploration target $t_i \in T$, we retrieve corresponding code content $c_i$ from $\mathcal{R}$ using $G(\mathcal{R})$ for efficient lookup. Each code element undergoes structured analysis:

\begin{equation}
    i_i = \text{LLM}_{analyze}(P_0, c_i, t_i)
\end{equation}

where the analysis extracts specific insight categories: role assessment describing how $c_i$ relates to the reported issue, fix location hints identifying potential modification points with confidence estimates, technical insights capturing implementation patterns and architectural decisions, and alternative hypotheses proposing different root cause explanations. The LLM analysis is also used to score the context for relevance, which we use to filter insights, retaining insights where $score_{{i_i}} \geq \tau_{rel}$, producing the filtered insight set $I_{filtered}$ used in the final synthesis stage.

\begin{figure*}
    \centering
    \includegraphics[width=0.8\linewidth]{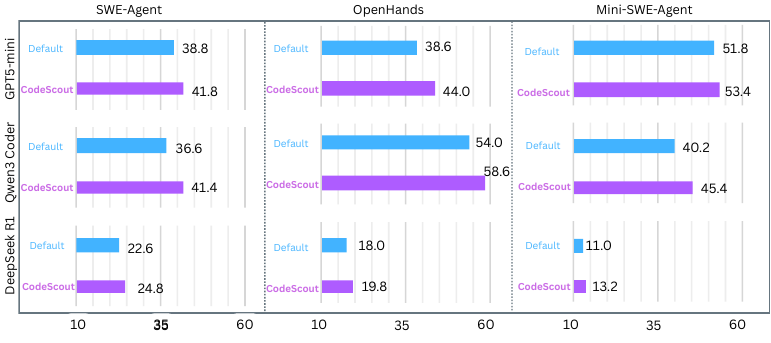}
    \caption{\small{Main comparison between default (no augmentation) and our contextual augmentation across three LLMs (GPT-5-mini, Qwen3 Coder, DeepSeek R1) and three scaffolds (SWE-Agent, OpenHands, Mini-SWE-Agent). Augmentation yields consistent improvements in resolution rate; gains are largest when the runtime LLM has weaker agentic abilities.}}
    \label{fig:main_results}
\end{figure*}

\subsection{Problem Synthesis}

The final synthesis stage generates the augmented specification through:

\begin{equation}
   P_{aug} = \mathcal{T}(P_0, \mathcal{R}, \mathcal{S})
\end{equation}

where $\mathcal{S} = \{G(\mathcal{R}), T, I_{filtered}\}$ represents the structured contextual knowledge extracted through the pipeline. In practice, the synthesis transformation operates by using an LLM to combine the original problem statement $P_0$ with the filtered insights $I_{filtered}$, as these insights already encapsulate the relevant information extracted from the repository through the knowledge graph and target selection stages.

The synthesis prompt structures the output as $P_{aug}$, which contains an enhanced description, reproduction steps, expected behavior specification, exploration hints, and fix guidance, where each component serves specific debugging functions.

Enhanced Issue Description integrates technical insights from the filtered analysis to clarify problem mechanisms and scope. Reproduction Steps augment original reproduction procedures with internal technical details and error patterns discovered during code analysis.
Exploration Hints map directly from analysis insights to provide structured guidance. Each hint includes specific reasoning derived from role analysis and technical insights components of relevant code elements.
Fix Hints synthesize fix location suggestions with confidence estimates. High-confidence locations are derived from the fix hint analysis with associated confidence scores, while alternative hypotheses aggregate insights across analyzed elements.

The synthesis process maintains traceability by preserving relationships between generated guidance and source code analysis. The resulting augmented specification $P_{aug}$ serves as a comprehensive debugging guide that significantly reduces initial investigation overhead compared to the original problem statement $P_0$.
\section{Experimental Setup}

We evaluate our approach on the SWEBench-Verified benchmark \cite{jimenez2023swe}, a widely used dataset software engineering agents on real-world issue-solving tasks.

Our experiments consider three scaffolds: SWE-agent \cite{yang2024sweagent}, OpenHands \cite{wang2025openhands}, and Mini-SWE-Agent. Each scaffold interfaces with different Large Language Models (LLMs), specifically GPT-5-mini, DeepSeek R1, and Qwen3 Coder 480B.

For \textbf{SWE-agent}, we enforce cost limits of 1.0, 1.0, and 0.75 USD per model, respectively. For \textbf{OpenHands}, we restrict execution to a maximum of 80 steps. All other configurations for both SWE-agent and OpenHands, as well as for \textbf{Mini-SWE-Agent}, follow their default settings.

\section{Results}

\subsection{Quantitative gains and localization}
We evaluate the effect of contextual problem-statement augmentation across three agentic scaffolds (SWE-Agent, OpenHands, Mini-SWE-Agent) and three LLM families (DeepSeek R1, Qwen3 Coder, GPT-5-mini). Across the full SWEBench-Verified evaluation the augmentation consistently improves the agent resolution rate: the gains are broadly consistent across scaffolds and models and are most pronounced when the runtime agent is relatively weak. Figure~\ref{fig:main_results} summarizes the aggregate performance gap between the default (no augmentation) and CodeScout across all tested combinations.

On the SWE-Agent subset used for ablation (Table~\ref{tab:ablations}) our full workflow increases the number of resolved issues compared to the Default baseline by absolute counts of \{+11, +15, +24\} for \{DeepSeek R1, GPT-5-mini, Qwen3 Coder\} respectively, corresponding to relative increases of approximately \{9.6\%, 7.7\%, 13.1\%\}. \sh{If people will ask why the absolute number is small, we can just highlight the percentage of the improvement.} These numbers underscore that the augmentation pipeline consistently converts underspecified problem statements into more actionable specifications that downstream agents can act upon.

We also evaluate \emph{localization} (whether the generated patch touches the same file or function as the ground-truth patch). At both file- and function-level granularity the augmented setting improves localization accuracy over the default across models. The improvement is particularly large for DeepSeek R1 (which lacks strong agentic reasoning out-of-the-box), indicating that workflow-driven augmentation helps weaker agents find the correct code regions. For GPT-5-mini, however, static localization metrics underestimate actual runtime localization: GPT-5-mini often produces patches that are applied at runtime (dynamic edits or runtime instrumentation), so static file-level matches do not fully capture its localization ability.

\begin{figure}
    \centering
    \includegraphics[width=\linewidth]{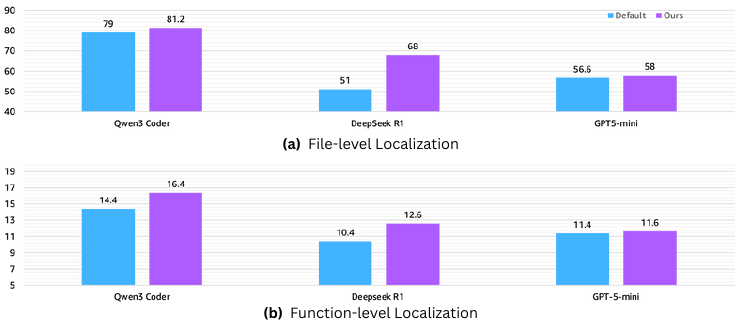}
    \caption{\small{Localization comparison (file-level and function-level) using SWE-Agent as the scaffold.} 
    }
    \label{fig:localization}
\end{figure}

\subsection{Ablation Study}
We performed ablations with SWE-Agent as the base scaffold to isolate which components of the workflow drive the gains (Table~\ref{tab:ablations}). Key observations:

\noindent\textbf{Does the full pipeline improve performance?} Yes, consistently across all models: DeepSeek R1 +11 resolved (+9.6\%), GPT-5-mini +15 (+7.7\%), Qwen3 Coder +24 (+13.1\%).

\noindent\textbf{Can agents autonomously perform augmentation?} When we modify system prompts to request autonomous augmentation during execution, performance drops significantly below Default for all three models (DeepSeek R1: -5, GPT-5-mini: -17, Qwen3 Coder: -25). This validates that CodeScout's separate, structured augmentation stage addresses a real limitation—agents cannot effectively self-augment during their trajectory.

\noindent\textbf{Is relevance filtering necessary?} Yes. Removing the relevance filter (\emph{CodeScout - without filtering}) reduces or eliminates much of the benefit (small or negative changes versus Default for some models), showing that careful filtering is crucial to avoid adding noisy context.

\noindent\textbf{Does LLM scoping outperform retrieval?} Yes. Replacing LLM-driven scoping with BM25 retrieval \cite{Bm25}—a classical lexical matching algorithm that ranks documents by term frequency—still improves over Default but yields smaller gains than LLM scoping, particularly for Qwen3 Coder. This suggests LLM scoping captures semantic relationships beyond lexical overlap.


\begin{table}[]
\centering
\resizebox{\columnwidth}{!}{%
\begin{tabular}{@{}lrrr@{}}
\toprule
\textbf{Ablations/Baselines} & \multicolumn{1}{l}{\textbf{DeepSeek R1}} & \multicolumn{1}{l}{\textbf{GPT-5-mini}} & \multicolumn{1}{l}{\textbf{Qwen3 Coder}} \\ \midrule
Default                       & 114          & 194          & 183          \\
\textbf{CodeScout (Ours)  }                        & \textbf{125} & \textbf{209} & \textbf{207} \\
Agentic Intra-trajectory Augmentation & 109          & 177          & 158          \\
CodeScout (No Filtering)      & 116          & 190          & 190          \\
CodeScout (BM25 Entity Selection)                   & 119          & 195          & 198          \\ \bottomrule
\end{tabular}%
}
\caption{\small{Ablation results (SWE-Agent). Numbers are counts of resolved issues on SWEBench-Verified.
}
}


\label{tab:ablations}
\end{table}

\subsection{Cost and token analysis}
\begin{figure}
    \centering
    \includegraphics[width=\linewidth]{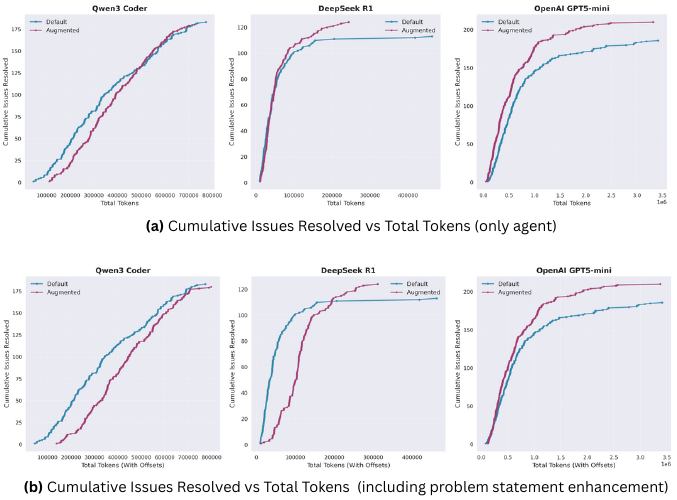}
    \caption{\small{Cumulative issues resolved as a function of total tokens consumed (input+output). (a) Agent-only token accounting; (b) token accounting including augmentation overhead. } 
    }
    
    \label{fig:token_vs_cumulative}
\end{figure}

\begin{figure}
    \centering
    \includegraphics[width=\linewidth]{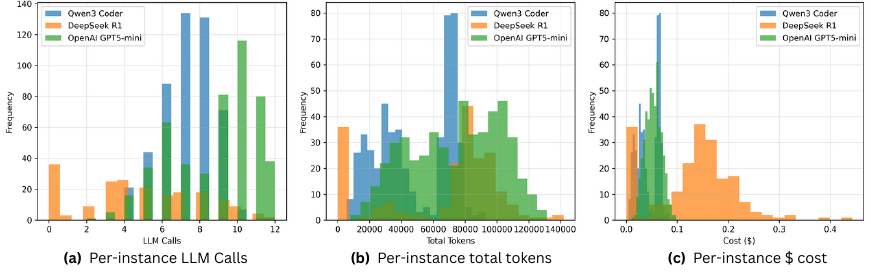}
    \caption{\small{Augmentation metrics: number of LLM calls, tokens and dollar cost per instance.}
    }
    \label{fig:aug_metrics}
\end{figure}

Figure~\ref{fig:token_vs_cumulative} and Figure~\ref{fig:aug_metrics} summarize token consumption and dollar cost trade-offs. \noindent\textbf{Tokens-per-resolved-issue improves for most models:} When comparing cumulative issues resolved at a given token budget (Fig.~\ref{fig:token_vs_cumulative}a), augmented runs resolve more issues for the same token budget for Qwen3 and DeepSeek. When we explicitly offset agent token consumption by the augmentation overhead (Fig.~\ref{fig:token_vs_cumulative}b), the gains remain comparable for Qwen3 and DeepSeek, showing that augmentation does not simply shift cost to a separate preprocessing step for these models. \noindent\textbf{Large-trajectory LLMs are an exception:} GPT-5-mini produces extremely long trajectories in our setup (order-of-magnitude more tokens per trajectory), so augmentation overhead is comparatively small and does not meaningfully change the tokens-per-resolved-issue metric. The net effect is that augmentation improves performance but the large absolute token budgets required by GPT-5-mini remain a deployment consideration.


\subsection{Impact on problem statements}
\begin{figure}
    \centering
    \includegraphics[width=\linewidth]{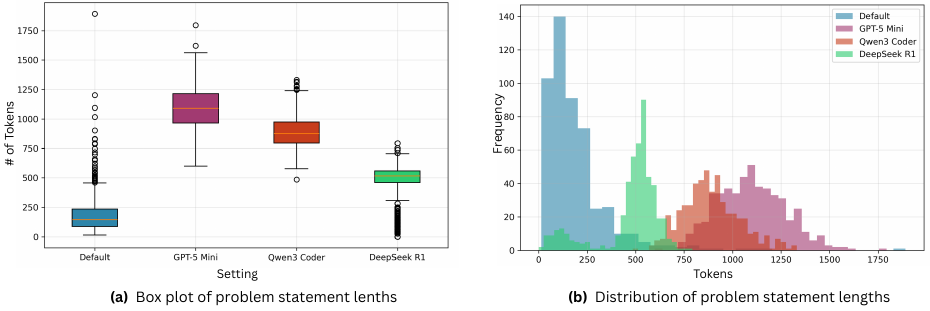}
    \caption{\small{Distribution of problem-statement lengths (token count) for Default vs Augmented. }
    }
    \label{fig:aug_len_dist}
\end{figure}

Augmentation increases the mean length of problem statements (Figure~\ref{fig:aug_len_dist}) but reduces noise in the long tail: default problem statements frequently include irrelevant large logs or environment dumps, while augmented statements are longer but more focused and actionable. The length-distribution plot shows a tighter, more concentrated augmented distribution and a reduced long tail relative to Default, indicating the augmentation mostly adds concentrated, high-value technical context rather than indiscriminate verbosity.

\begin{figure*}[h]
    \centering
    \includegraphics[width=\linewidth]{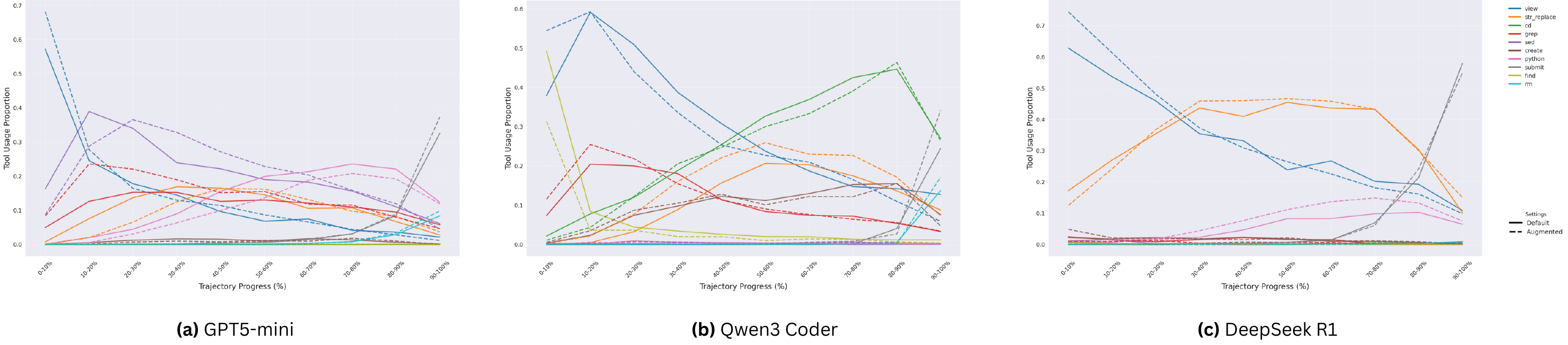}
    \caption{\small{Evolution of tool-call proportions across a trajectory (Default vs CodeScout) for SWE-Agent. }
    }
    \label{fig:tool_calls}
\end{figure*}
\subsection{Agent behavior and tool usage}
Trajectory analysis (Figure~\ref{fig:tool_calls}) shows a systematic change in early-agent behavior under augmentation: the initial fraction of calls to \texttt{view} and \texttt{grep} rises and \texttt{find} falls, consistent with more targeted exploration. Later phases show expected increases in active repository actions (patching, creates, runs). These changes indicate that augmentation enables agents to begin trajectories with higher-quality search terms and file targets, reducing wasted broad exploration.

\subsection{Cross-synthesis}
Table~\ref{tab:cross_synthesis} (cross-synthesis) evaluates combinations where the model used to augment problem statements differs from the model used to run the agent. The key pattern is asymmetry in gains:

\noindent\textbf{Weak runtime LLMs benefit most from stronger augmenters:} DeepSeek R1 as the runtime model sees its largest improvement when \emph{augmented} by Qwen3 Coder: from 108 to 164 resolved issues (+56, +51.9\%). This demonstrates that a stronger off-line augmenter can compensate for limited runtime agent capabilities.

\noindent\textbf{Strong runtime LLMs gain modestly from weaker augmenters:} When GPT-5-mini runs the agent, augmenting with weaker models (DeepSeek R1) still yields small improvements (194 to 196, +2, +1.0\%), showing robustness but diminishing returns.

\noindent\textbf{Practical implication:} one can use a cheaper, capable model to pre-compute augmentations for many instances (amortizing cost) and still obtain material improvements for weaker runtime agents; conversely, using a stronger augmenter can substantially raise the performance floor for weak agents.


\begin{table}[]
\centering
\resizebox{\columnwidth}{!}{%
\begin{tabular}{@{}l|r|rrr@{}}
\toprule
\multirow{2}{*}{Agent Base LLM} &
  \multicolumn{1}{c|}{\multirow{2}{*}{Default}} &
  \multicolumn{3}{c}{Augmented by} \\ \cmidrule(l){3-5} 
 &
  \multicolumn{1}{c|}{} &
  \multicolumn{1}{l}{DeepSeek R1} &
  \multicolumn{1}{l}{Qwen3 Coder} &
  \multicolumn{1}{l}{GPT5-mini} \\ \midrule
DeepSeek R1 & 108 & 125 & 164 & 132 \\
Qwen3 Coder & 183 & 194 & 209 & 190 \\
GPT5-mini   & 194 & 196 & 207 & 209 \\ \bottomrule
\end{tabular}%
}
\caption{\small{Cross-synthesis results: rows are the base (runtime) model used by the agent; columns show resolved counts when problem-statements are augmented by the indicated model. }}
\label{tab:cross_synthesis}
\end{table}

\subsection{Takeaways}
\sh{Add a summary at the beginning.} \ms{Yes.}
\xiangci{Highlight some of these points in abstract and introduction as well. These are very strong results.}
\xiangci{From this section, the rest part more like AI-generated somehow. I can feel the wording and style difference. Can you please rewrite and smooth it.}
Our evaluation demonstrates that CodeScout consistently enhances agent performance across diverse models and scaffolds, with structured augmentation outperforming self-augmentation approaches. These findings validate our hypothesis that systematic pre-exploration effectively supplements existing agent architectures.
\noindent \textbf{Effectiveness:} Contextual problem-statement augmentation consistently improves resolution rates and localization across scaffolds and LLMs, with the largest relative gains for weaker runtime agents.

\noindent \textbf{Pipeline vs autonomous augmentation:} A separate, structured augmentation pipeline outperforms asking the agent to self-augment during its trajectory.

\noindent \textbf{Design trade-offs:} LLM-driven scoping plus relevance filtering is important—simple lexical retrieval (BM25) helps, but LLM scoping yields stronger, more semantically-relevant targets. Augmentation overhead (LLM calls, tokens) is modest relative to downstream gains for most models, though very large-trajectory LLMs maintain higher absolute token budgets.

\noindent \textbf{Deployment guidance:} For production settings consider (a) using a moderately capable, cheaper model for augmentation to boost weaker runtime agents, or (b) using a stronger augmenter to raise the floor of weaker agents when improving the runtime model is not feasible.



\section{Conclusion}
We introduced a contextual query refinement framework that transforms vague and underspecified user requests into detailed, actionable problem statements through structured codebase analysis. By combining repository graph construction, high-level context scoping, and multi-perspective code element analysis, our approach systematically grounds problem understanding before attempting solutions. Evaluation on the SWEBench-Verified benchmark demonstrates that this refinement process enhances the effectiveness of AI code assistance, particularly when operating in poorly defined scenarios. Beyond performance gains, our findings highlight the importance of structured problem formulation as a prerequisite for reliable software engineering support. Future work will explore extending this framework to multi-repository settings, integrating deeper program analysis techniques, and applying the approach to collaborative development environments.

\xiangci{Don't forget to add the mandatory Limiations and Ethics Statement section!}

\section{Limitations}

Our evaluation is necessarily constrained by the current state of software engineering benchmarks and tooling. We evaluate CodeScout using three language models (DeepSeek R1, GPT-5-mini, Qwen3 Coder) and three agentic scaffolds (SWE-Agent, OpenHands, Mini-SWE-Agent), which, while representing state-of-the-art systems, constitute only a subset of available approaches. Expanding to additional models and scaffolds would strengthen claims about generalizability but requires substantial computational resources and careful experimental design to ensure fair comparison. Similarly, our analysis focuses exclusively on Python repositories, not by choice but due to the availability of established benchmarks like SWEBench-Verified with reliable evaluation harnesses. Other programming languages lack comparable evaluation infrastructure that provides ground-truth issue resolution verification at scale, preventing systematic assessment of CodeScout's effectiveness across language ecosystems with different structural conventions, type systems, and semantic properties. Finally, our evaluation relies on open-source repositories from SWEBench-Verified, which may differ systematically from enterprise codebases in terms of architectural patterns, documentation practices, code quality standards, and proprietary constraints. While open-source repositories enable reproducible research, validating CodeScout's effectiveness on closed-source or domain-specific industrial code remains important future work that requires industry partnerships and appropriate data sharing agreements.

\section{Ethics Statement}

Our work focuses on improving AI-assisted software development through enhanced problem statement formulation. We acknowledge the inherent risks of LLM-based code assistance systems, including potential generation of insecure code, introduction of subtle bugs, and perpetuation of biases present in training data. While CodeScout aims to improve agent reliability, users should apply appropriate code review and security practices to any AI-generated solutions. All datasets used in this work—including SWEBench-Verified and associated repositories—are used in accordance with their respective licenses and usage terms. We do not introduce new datasets or collect proprietary code. Our evaluation respects the open-source nature of the benchmark and does not compromise repository security or developer privacy. The computational resources required for our experiments contribute to environmental impact through energy consumption. We encourage practitioners to consider this tradeoff when deploying augmentation systems at scale.

\appendix

\appendixpage

\startcontents[sections]
\printcontents[sections]{l}{1}{\setcounter{tocdepth}{3}}

\section{Prompt Design}
\label{app:prompts}

We employ a three-stage prompting strategy to systematically augment problem statements with codebase context. Each prompt is designed to decompose the augmentation task into focused subtasks that elicit specific types of knowledge from the language model. Below we present the complete prompt templates.

\subsection{Exploration Target Identification}
\label{app:prompt-explore}

The first prompt guides the model to identify relevant code locations given a problem statement and repository structure. This stage establishes which parts of the codebase warrant deeper investigation.

\noindent\textbf{Input:} Problem statement $\text{PS}$, repository file tree $\mathcal{F}$

\noindent\textbf{Output:} List of exploration targets $\mathcal{T} = \{(\text{type}, \text{name}, \text{reasoning})\}$

\begin{tcolorbox}[colback=gray!5!white,colframe=gray!75!black,title=Stage 1: Exploration Target Identification]
\small
You are analyzing a software bug report to identify which files and code components to investigate.

\vspace{0.3em}
\textbf{PROBLEM STATEMENT:}

\texttt{\{problem\_statement\}}

\vspace{0.3em}
\textbf{REPOSITORY STRUCTURE:}

\texttt{\{file\_tree\}}

\vspace{0.3em}
\textbf{Task:} Identify the most promising files, classes, and functions to explore for understanding this issue.

\vspace{0.3em}
\textbf{Focus on:}
\begin{itemize}
    \item Files/classes mentioned in the problem statement
    \item Related components that might be affected
    \item Common entry points and core functionality
    \item Test files that might reveal expected behavior
\end{itemize}

\vspace{0.3em}
\textbf{Return exactly 5--10 targets in this format:}

\texttt{1. target\_type:file, target\_name:exact\_name, reasoning:why\_relevant}

\texttt{2. target\_type:class, target\_name:exact\_name, reasoning:why\_relevant}

\texttt{3. target\_type:function, target\_name:exact\_name, reasoning:why\_relevant}

\vspace{0.3em}
Be specific with names---use exact file paths and class/function names from the tree. Each entry must be on a separate line and follow the exact format shown.
\end{tcolorbox}

\noindent\textbf{Design rationale:} We ask the model to produce 5--10 targets to balance coverage with computational cost. The structured output format enables reliable parsing. By requesting explicit reasoning, we encourage the model to articulate its hypotheses about bug locations, which improves target quality.

\subsection{Content Relevance Analysis}
\label{app:prompt-analyze}

For each exploration target, we retrieve its source code and prompt the model to analyze how it relates to the reported issue. This stage extracts structured insights that will inform the final augmented problem statement.

\noindent\textbf{Input:} Problem statement $\text{PS}$, code content $c$, target metadata $\theta$

\noindent\textbf{Output:} Structured analysis with relevance score and insights

\begin{tcolorbox}[colback=gray!5!white,colframe=gray!75!black,title=Stage 2: Content Relevance Analysis]
\small
You are analyzing code content to understand how it relates to a specific bug report. Your analysis will directly feed into an augmented problem statement.

\vspace{0.3em}
\textbf{ORIGINAL PROBLEM:}

\texttt{\{problem\_statement\}}

\vspace{0.3em}
\textbf{CODE CONTENT:}

\texttt{\{content\}}

\vspace{0.3em}
\textbf{TARGET INFO:} \texttt{\{target.target\_type\} - \{target.target\_name\}}

\vspace{0.3em}
\textbf{Provide analysis in the following format:}

\vspace{0.2em}
\textbf{RELEVANCE:} X/10

\vspace{0.2em}
\textbf{ROLE\_IN\_ISSUE:}

Describe how this code relates to the reported issue. What is its role in causing or manifesting the bug?

\vspace{0.2em}
\textbf{FIX\_LOCATION\_HINTS:}

Identify specific locations where fixes might be needed (line numbers, methods, conditions). What type of changes might be required (validation, error handling, logic fixes)?

\vspace{0.2em}
\textbf{EXPLORATION\_INSIGHTS:}

Surface key technical details that help understand the issue---algorithms, data flow, dependencies, code patterns, or architectural decisions relevant to the bug. Note test scenarios or edge cases this code suggests should be examined.

\vspace{0.2em}
\textbf{ALTERNATIVE\_HYPOTHESES:}

Consider alternative explanations for the bug based on this code. What related components might also be involved? What different root causes does this analysis suggest?

\vspace{0.3em}
\textbf{Note:} Focus on insights that will help create actionable exploration hints and fix suggestions. If relevance $<$ 4, keep all sections brief but still provide analysis.
\end{tcolorbox}

\noindent\textbf{Design rationale:} The structured output schema aligns with the final augmented problem statement format. By requesting a numerical relevance score, we enable downstream filtering of low-relevance content. The four semantic categories (role, fix hints, insights, hypotheses) encourage comprehensive analysis while maintaining organization.

\subsection{Problem Statement Synthesis}
\label{app:prompt-synthesize}

Given the original problem statement and all relevant code analyses, we prompt the model to synthesize a comprehensive augmented problem statement. This stage transforms raw insights into a coherent narrative that guides developers.

\noindent\textbf{Input:} Original problem statement $\text{PS}$, filtered analyses $\mathcal{A}_{\text{rel}}$

\noindent\textbf{Output:} Augmented problem statement $\text{PS}'$

\begin{tcolorbox}[colback=gray!5!white,colframe=gray!75!black,title=Stage 3: Problem Statement Synthesis]

\small
You are creating an enhanced problem statement that will help developers quickly understand and fix a software issue.

\textbf{ORIGINAL PROBLEM STATEMENT:} \texttt{\{original\_ps\}}

\textbf{CODE ANALYSIS RESULTS:} For each analyzed target:
\begin{verbatim}
{i}. TARGET: {target_type} - {target_name}
   RELEVANCE: {score}/10
   ROLE IN ISSUE: {role_description}
   FIX LOCATION HINTS: {hints}
   EXPLORATION INSIGHTS: {insights}
   ALTERNATIVE HYPOTHESES: {hypotheses}
\end{verbatim}

\textbf{Instructions:} Synthesize these analyses into a comprehensive problem statement. Include all relevant details from the original statement. Use the code analysis as context to inform your narrative---don't simply copy it verbatim.

\textbf{Structure your output with these sections:}

\textbf{\#\# Issue Description:} Rewrite the core problem clearly, incorporating technical insights.

\textbf{\#\# Reproduction Steps:} Include original steps if available. If the original issue includes code, preserve it here---this is essential. Add technical details about what happens internally. Include relevant stack traces or error patterns. Suggest additional reproduction scenarios if helpful.

\textbf{\#\# Expected Behavior:} Describe what should happen instead, informed by code analysis.

\textbf{\#\# Exploration Hints:}
\begin{itemize}[leftmargin=*,topsep=1pt,itemsep=0pt]
    \item \textit{Files to Examine:} \texttt{file1.py} - why relevant (role in issue, key findings)
    \item \textit{Key Classes/Functions:} \texttt{ClassName.method()} - what to look for
    \item \textit{Areas of Interest:} Specific code patterns or logic to investigate
\end{itemize}

\textbf{\#\# Fix Hints:}
\begin{itemize}[leftmargin=*,topsep=1pt,itemsep=0pt]
    \item \textit{High Confidence Locations:} Where fixes are likely needed and why
    \item \textit{Implementation Suggestions:} Concrete fix ideas with limitations
    \item \textit{Alternative Hypotheses:} Other potential root causes
\end{itemize}

Keep technical language precise but accessible. Focus on actionable insights.
\end{tcolorbox}

\noindent\textbf{Design rationale:} The detailed template structure ensures consistency across augmented problem statements. By explicitly requesting preservation of reproduction code and technical details, we maintain the diagnostic value of the original statement while enriching it with codebase context. The hierarchical organization mirrors how developers naturally think about debugging: understanding the issue, knowing where to look, and hypothesizing about fixes.

\section{Augmentation Algorithm}
\label{app:algorithm}

Our approach processes each problem statement through a structured pipeline that leverages language models to systematically gather and synthesize codebase context. Algorithm~\ref{alg:process-instance} presents the core augmentation procedure.

\subsection{Main Augmentation Procedure}

\begin{algorithm}[t]
\caption{Problem Statement Augmentation}
\label{alg:process-instance}
\begin{algorithmic}[1]
\STATE \textbf{Input:} Problem statement $\text{PS}$, codebase $\mathcal{C}$, language model $\mathcal{M}$, threshold $\tau$
\STATE \textbf{Output:} Augmented problem statement $\text{PS}'$
\STATE
\STATE // \textbf{Stage 1: Exploration Target Identification}
\STATE $\mathcal{F} \gets \textsc{GetFileTree}(\mathcal{C})$
\STATE $p_1 \gets \textsc{FormatPrompt}_{\text{explore}}(\text{PS}, \mathcal{F})$
\STATE $r_1 \gets \mathcal{M}(p_1)$
\STATE $\mathcal{T} \gets \textsc{ParseTargets}(r_1)$
\STATE
\STATE // \textbf{Stage 2: Content Retrieval and Analysis}
\STATE $\mathcal{A} \gets [\,]$
\FOR{each target $\theta \in \mathcal{T}$}
    \STATE $\text{content} \gets \textsc{RetrieveCode}(\theta, \mathcal{C})$
    \IF{content successfully retrieved}
        \STATE $p_2 \gets \textsc{FormatPrompt}_{\text{analyze}}(\text{PS}, \text{content}, \theta)$
        \STATE $r_2 \gets \mathcal{M}(p_2)$
        \STATE $a \gets \textsc{ParseAnalysis}(r_2)$
        \STATE $\mathcal{A} \gets \mathcal{A} \cup \{a\}$
    \ENDIF
\ENDFOR
\STATE
\STATE // \textbf{Stage 3: Relevance Filtering}
\STATE $\mathcal{A}_{\text{rel}} \gets \{a \in \mathcal{A} \mid a.\text{relevance} \geq \tau\}$
\STATE
\STATE // \textbf{Stage 4: Synthesis}
\STATE $\text{insights} \gets \textsc{FormatInsights}(\mathcal{A}_{\text{rel}})$
\STATE $p_3 \gets \textsc{FormatPrompt}_{\text{synthesize}}(\text{PS}, \text{insights})$
\STATE $\text{PS}' \gets \mathcal{M}(p_3)$
\STATE
\STATE \textbf{return} $\text{PS}'$
\end{algorithmic}
\end{algorithm}

\paragraph{Key design choices:} We employ a \textbf{relevance threshold} $\tau$ to filter analyses before synthesis, ensuring only high-quality insights inform the final augmented statement. The pipeline gracefully handles retrieval failures by skipping unavailable targets rather than halting, making it robust to incomplete code references.

\subsection{Computational Complexity}

\paragraph{LLM API Calls:} Each problem statement requires $2 + |\mathcal{T}_{\text{valid}}|$ language model calls:
\begin{itemize}
    \item 1 call for exploration target identification
    \item $|\mathcal{T}_{\text{valid}}|$ calls for content analysis (typically 5--10 targets)
    \item 1 call for synthesis
\end{itemize}

For a dataset of size $n$, the total number of LLM calls is $O(n \cdot |\mathcal{T}|)$. In practice, with approximately 7 valid targets per instance, we observe roughly 9 API calls per problem statement.

\section{CodeScout: Further Analysis}
\label{app:codescout_examples}

\subsection{Qualitative Examples}
We present detailed visualizations of the CodeScout context retrieval pipeline across five diverse instances from the SWE-bench Verified dataset. Each instance demonstrates the three-stage pipeline: (1) high-level context scoping to identify relevant files, classes, and functions, (2) fine-grained content analysis with relevance scoring, and (3) problem statement augmentation. We show results from three different retrieval methods using DeepSeek-R1, Qwen3-Coder, and GPT-5-mini.

\begin{figure*}
\centering
\includegraphics[width=\textwidth,page=1]{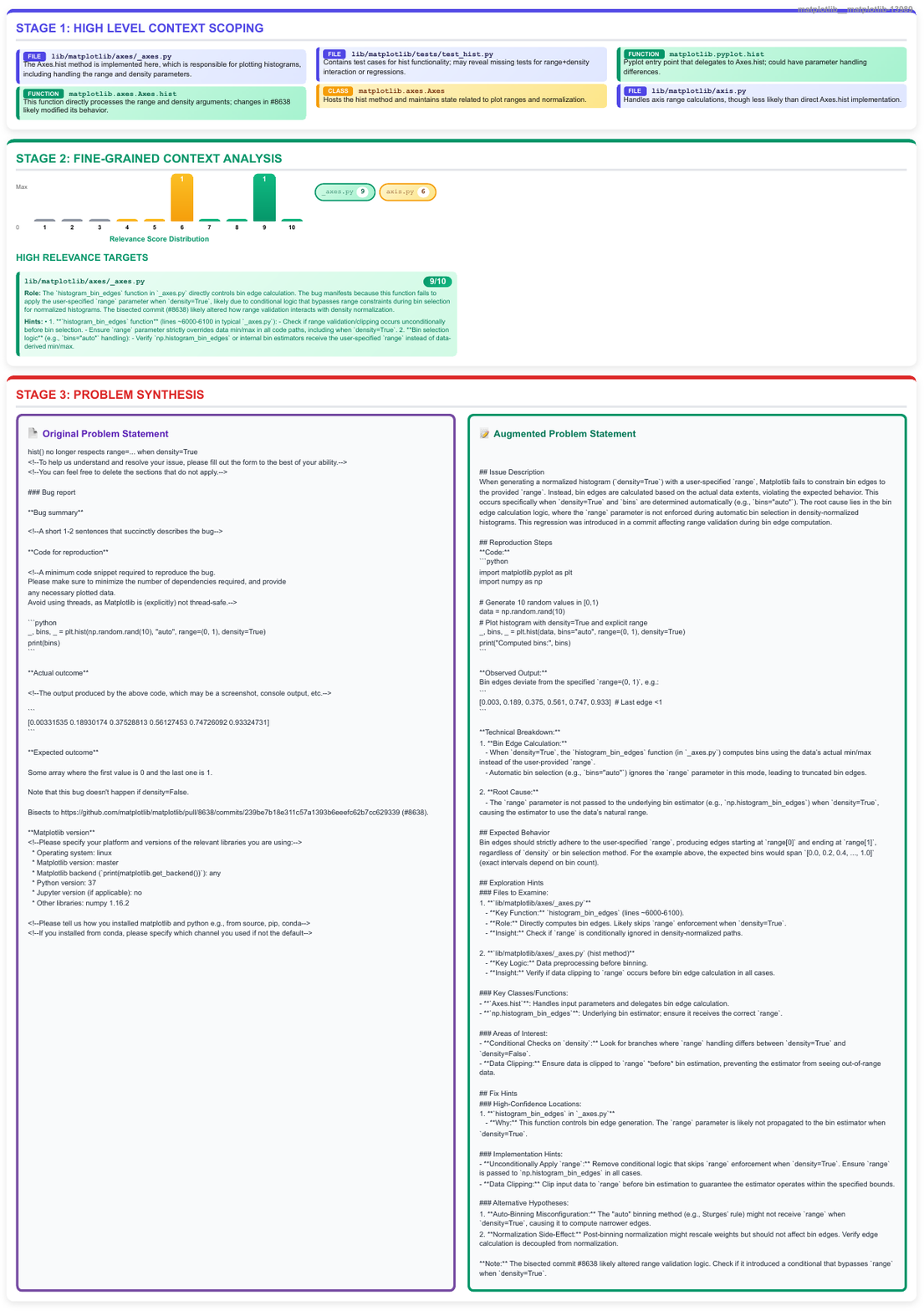}
\caption{CodeScout visualization for \texttt{matplotlib\_\_matplotlib-13989} with DeepSeek-R1.}
\label{fig:matplotlib-dsr1}
\end{figure*}

\begin{figure*}
\centering
\includegraphics[width=\textwidth,page=1]{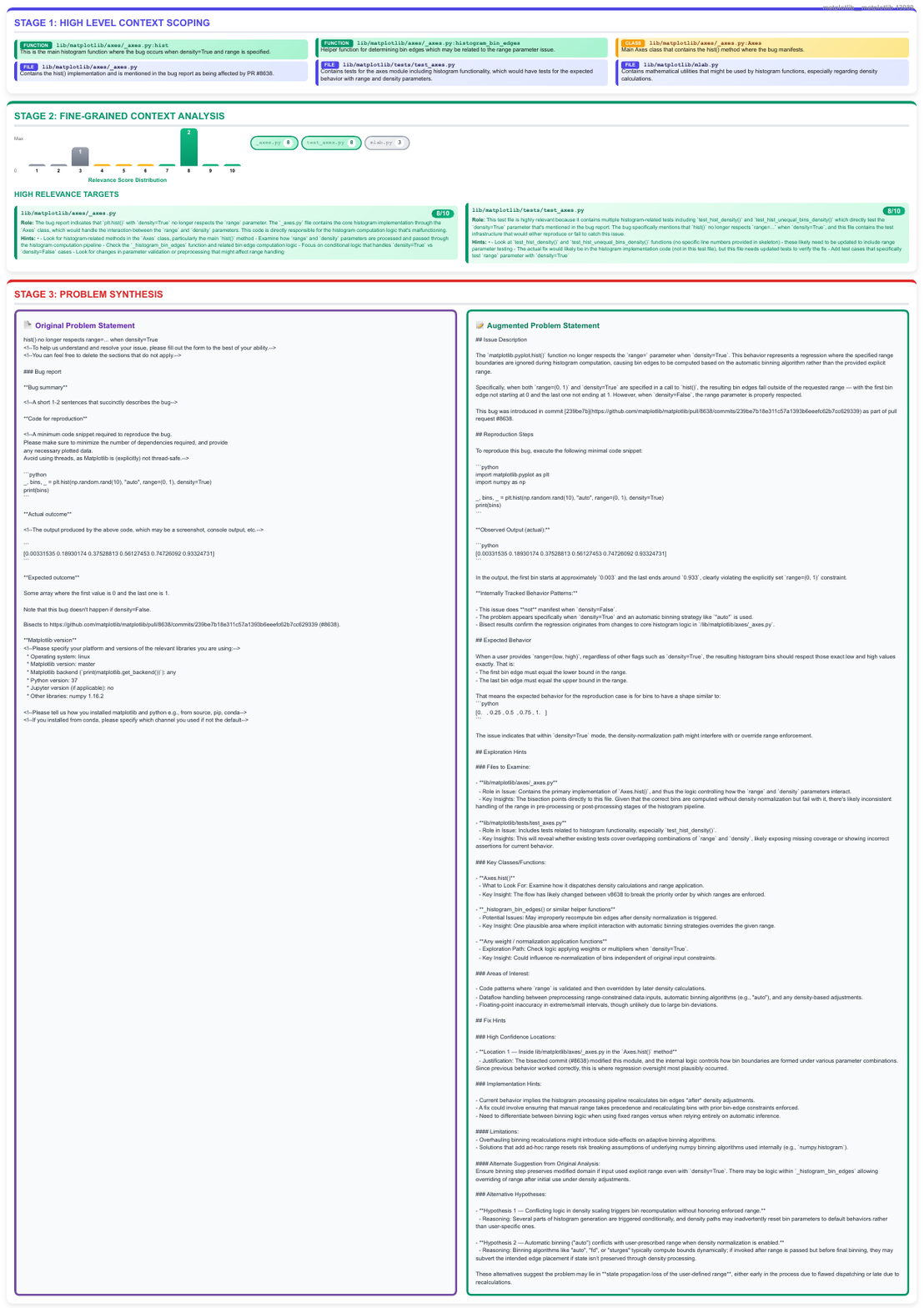}
\caption{CodeScout visualization for \texttt{matplotlib\_\_matplotlib-13989} with Qwen3-Coder.}
\label{fig:matplotlib-q3c}
\end{figure*}

\begin{figure*}
\centering
\includegraphics[width=\textwidth,page=1]{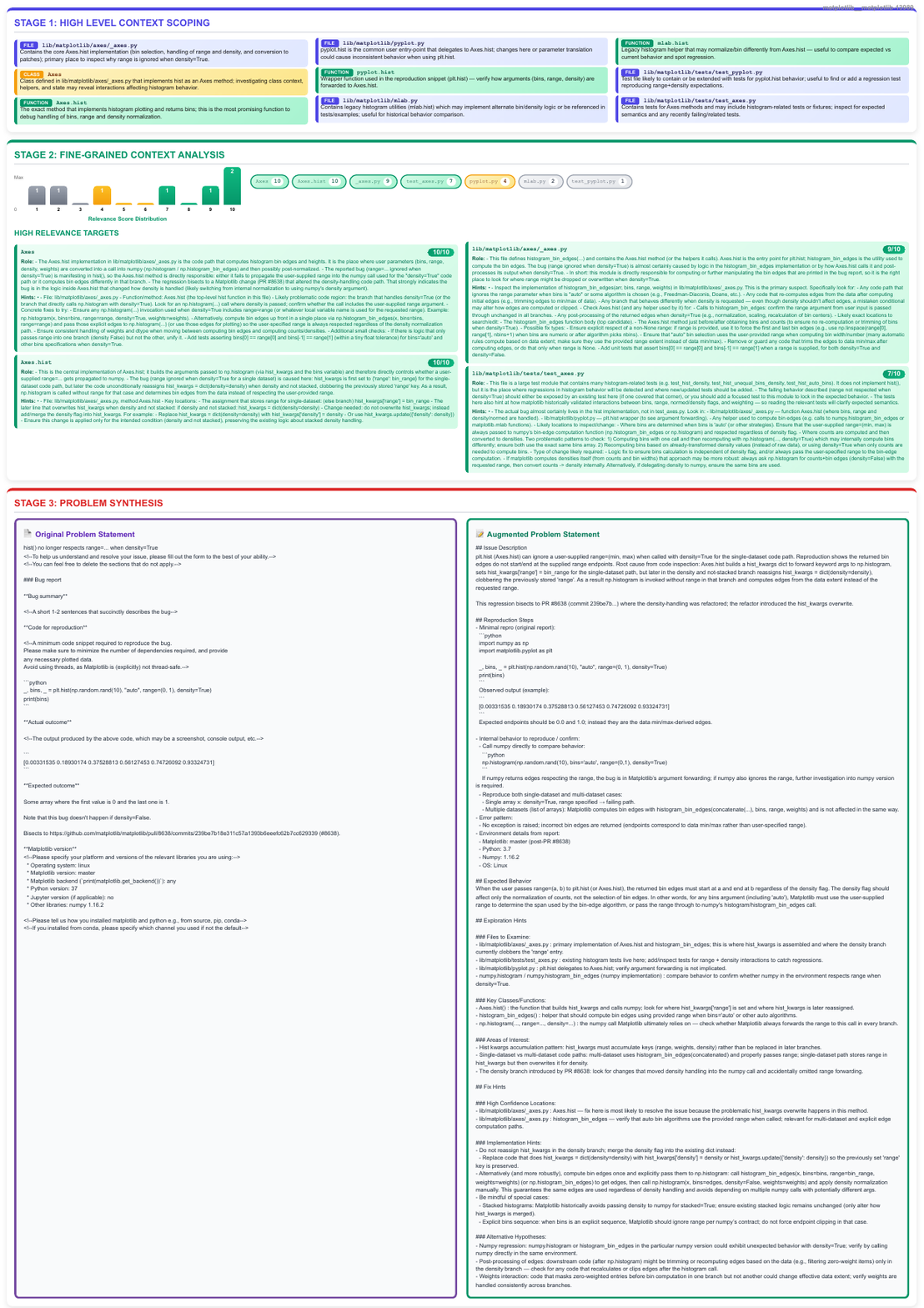}
\caption{CodeScout visualization for \texttt{matplotlib\_\_matplotlib-13989} with GPT-5-mini.}
\label{fig:matplotlib-gpt5}
\end{figure*}

\begin{figure*}
\centering
\includegraphics[width=\textwidth,page=1]{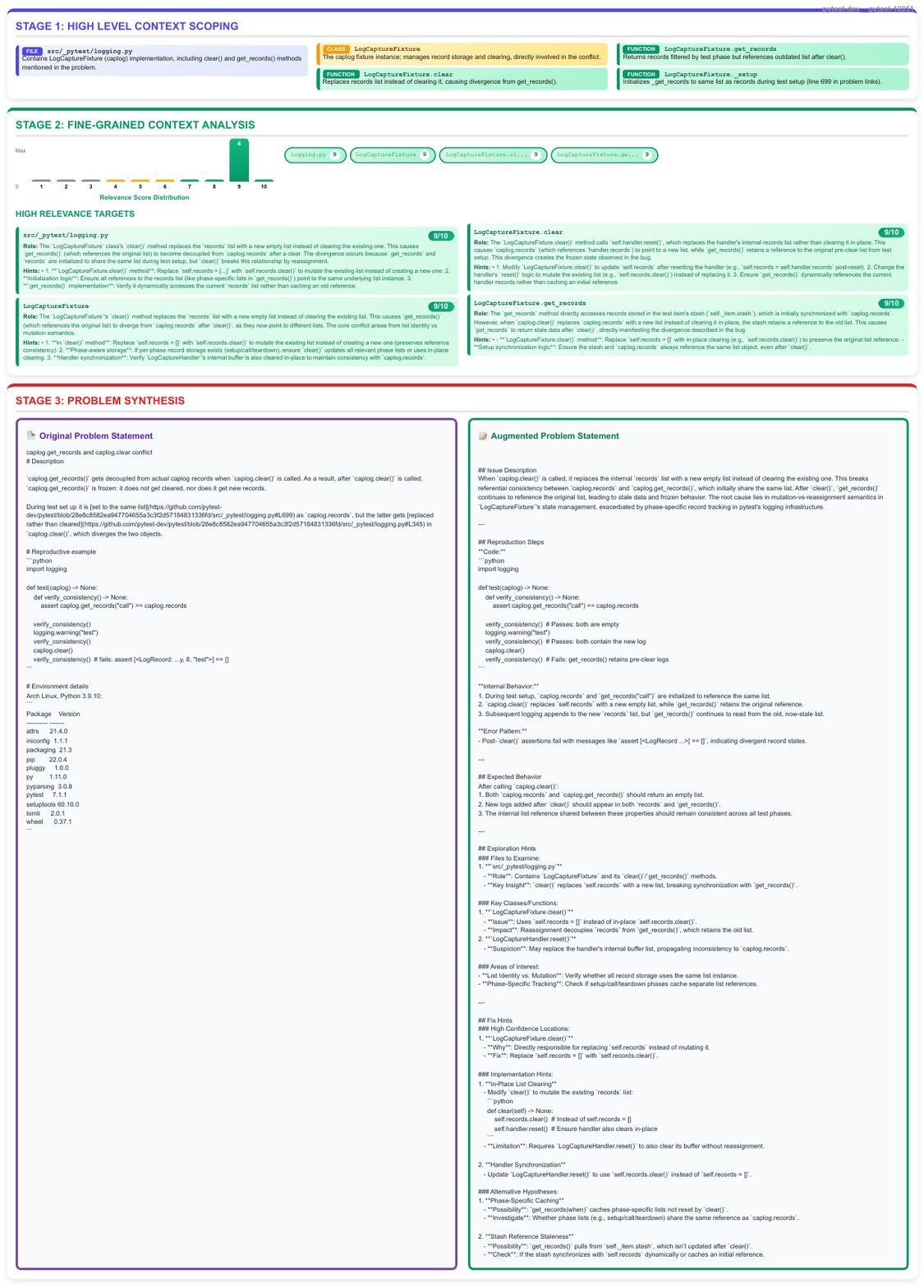}
\caption{CodeScout visualization for \texttt{pytest-dev\_\_pytest-10051} with DeepSeek-R1.}
\label{fig:pytest-dsr1}
\end{figure*}

\begin{figure*}
\centering
\includegraphics[width=\textwidth,page=1]{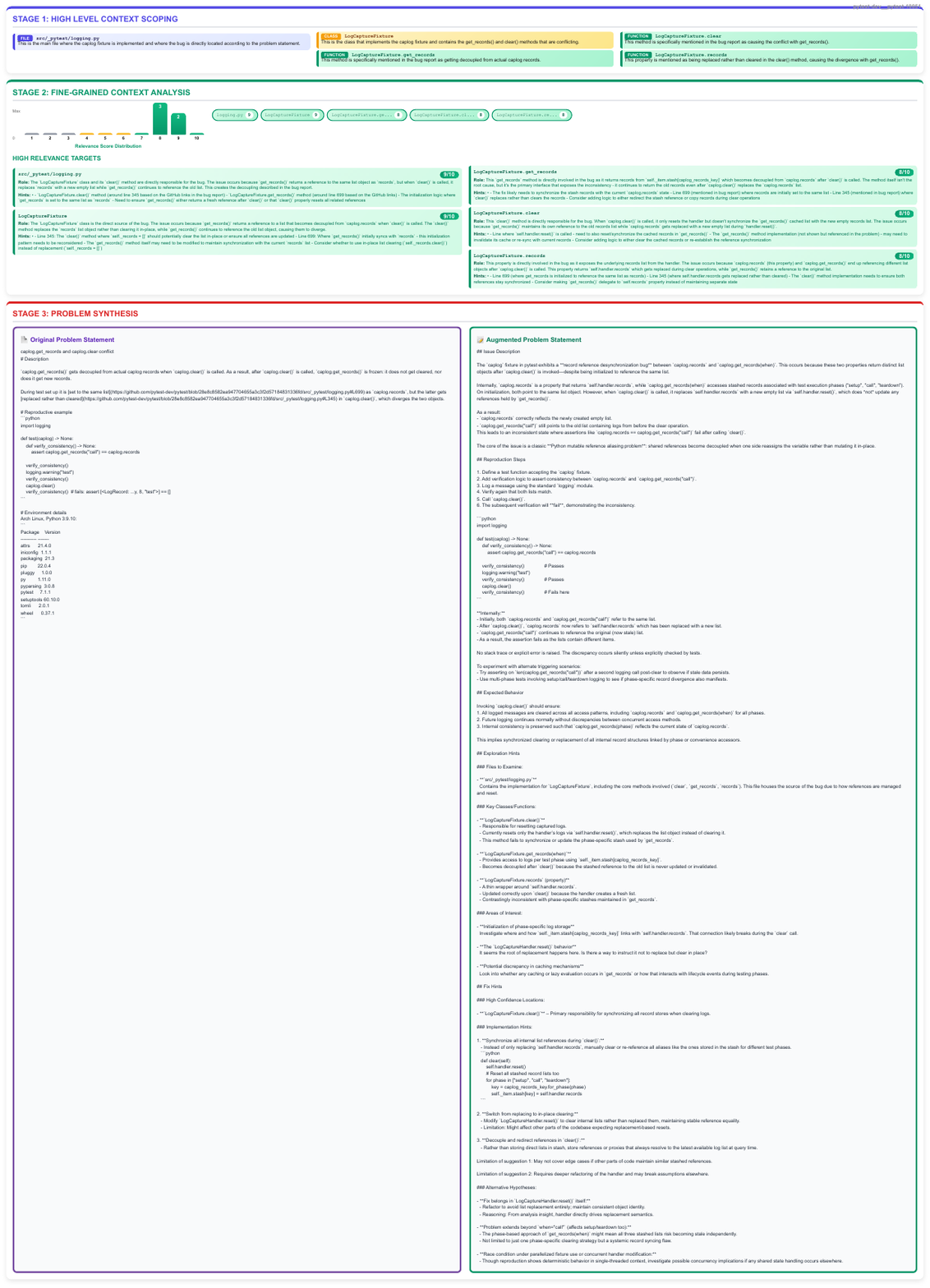}
\caption{CodeScout visualization for \texttt{pytest-dev\_\_pytest-10051} with Qwen3-Coder.}
\label{fig:pytest-q3c}
\end{figure*}

\begin{figure*}
\centering
\includegraphics[width=\textwidth,page=1]{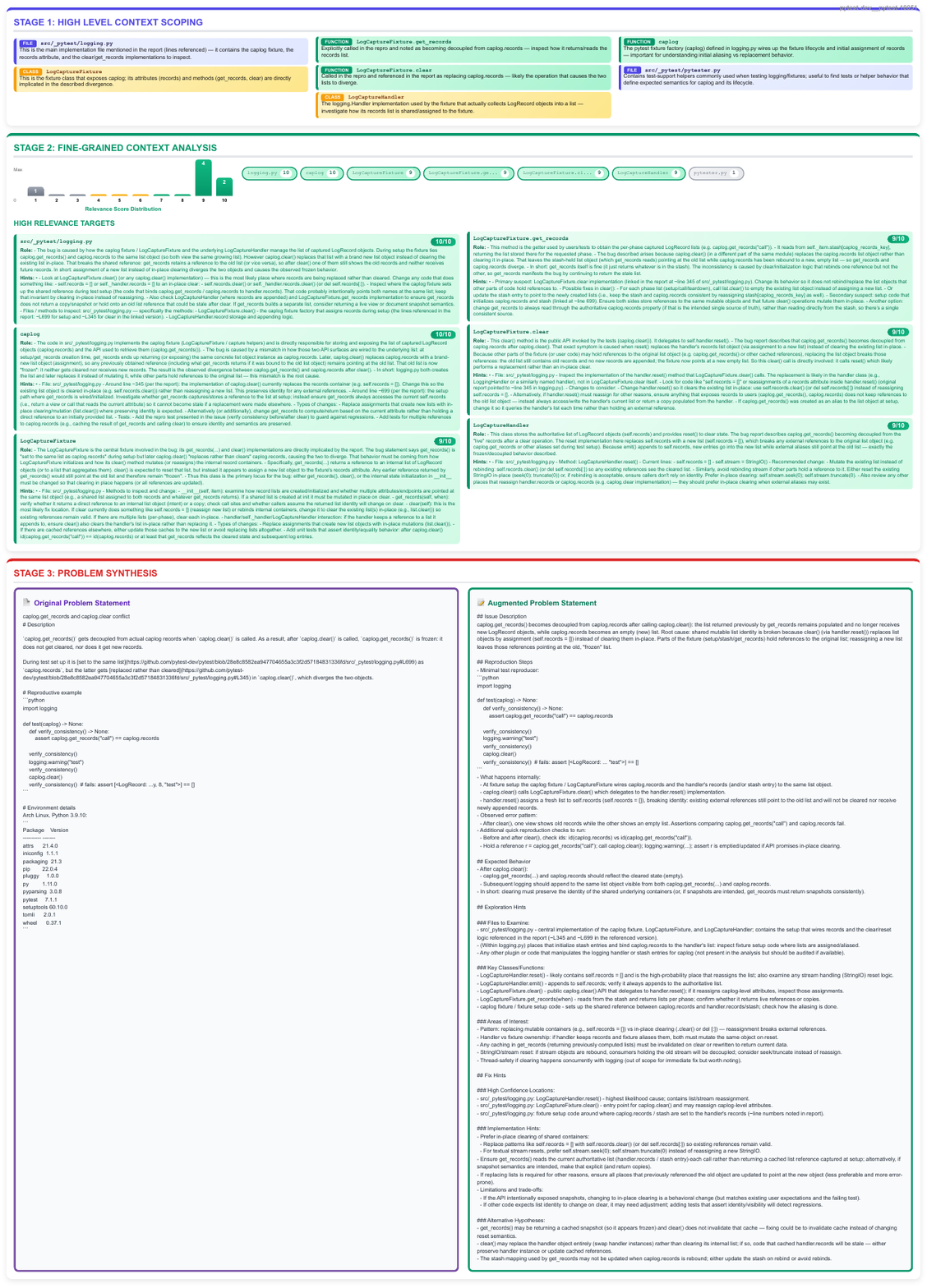}
\caption{CodeScout visualization for \texttt{pytest-dev\_\_pytest-10051} with GPT-5-mini.}
\label{fig:pytest-gpt5}
\end{figure*}

\begin{figure*}
\centering
\includegraphics[width=\textwidth,page=1]{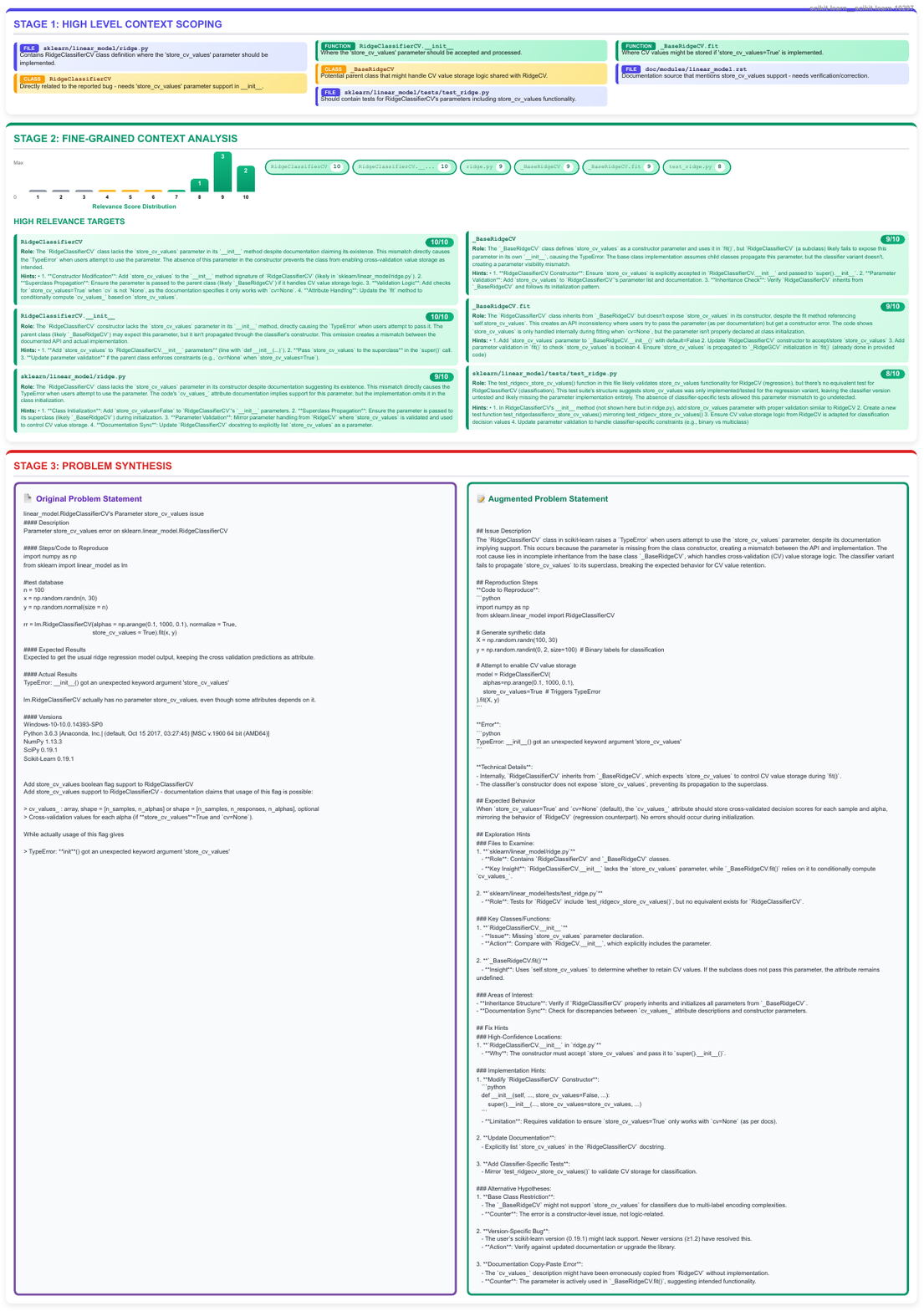}
\caption{CodeScout visualization for \texttt{scikit-learn\_\_scikit-learn-10297} with DeepSeek-R1.}
\label{fig:sklearn-dsr1}
\end{figure*}

\begin{figure*}
\centering
\includegraphics[width=\textwidth,page=1]{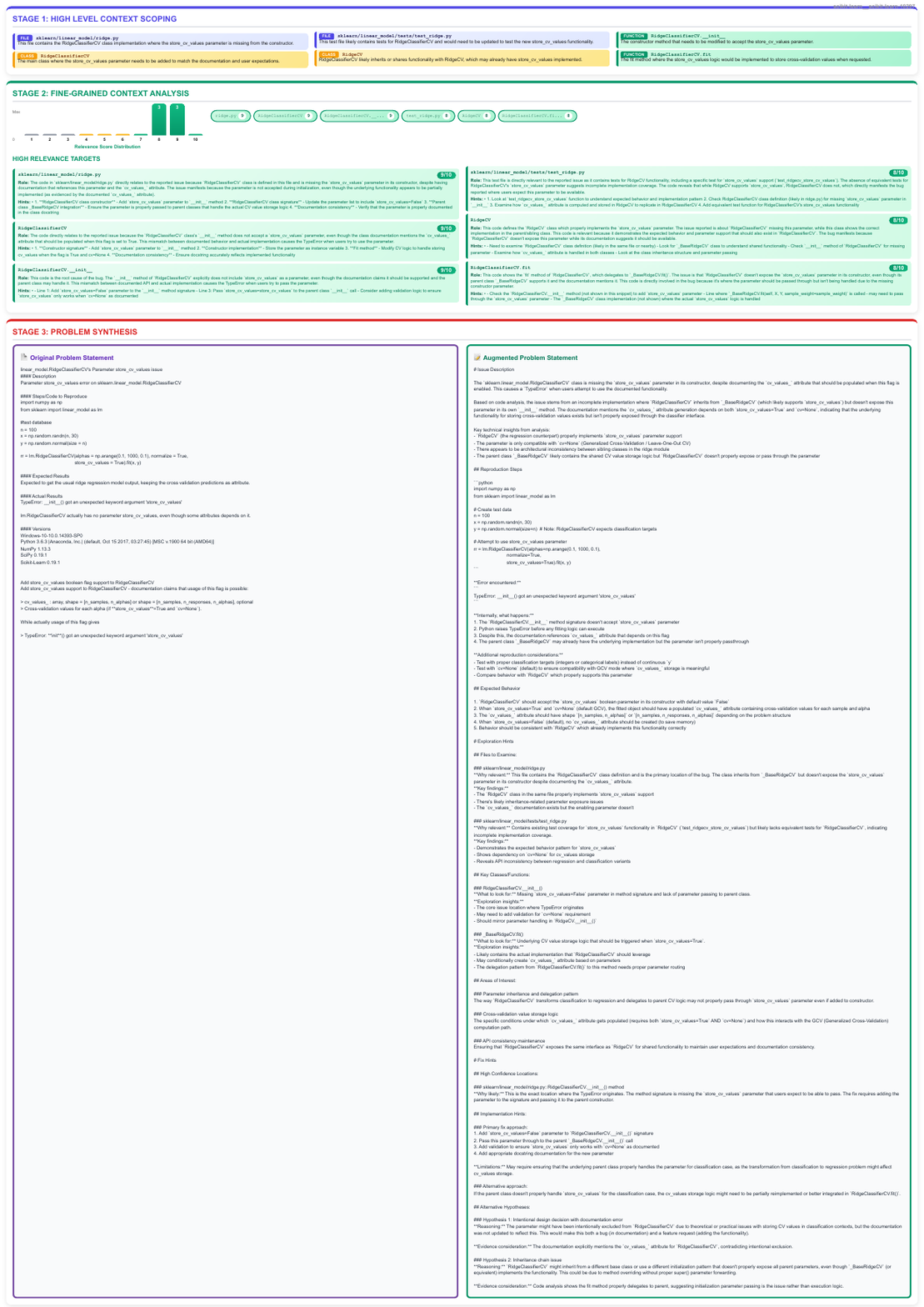}
\caption{CodeScout visualization for \texttt{scikit-learn\_\_scikit-learn-10297} with Qwen3-Coder.}
\label{fig:sklearn-q3c}
\end{figure*}

\begin{figure*}
\centering
\includegraphics[width=\textwidth,page=1]{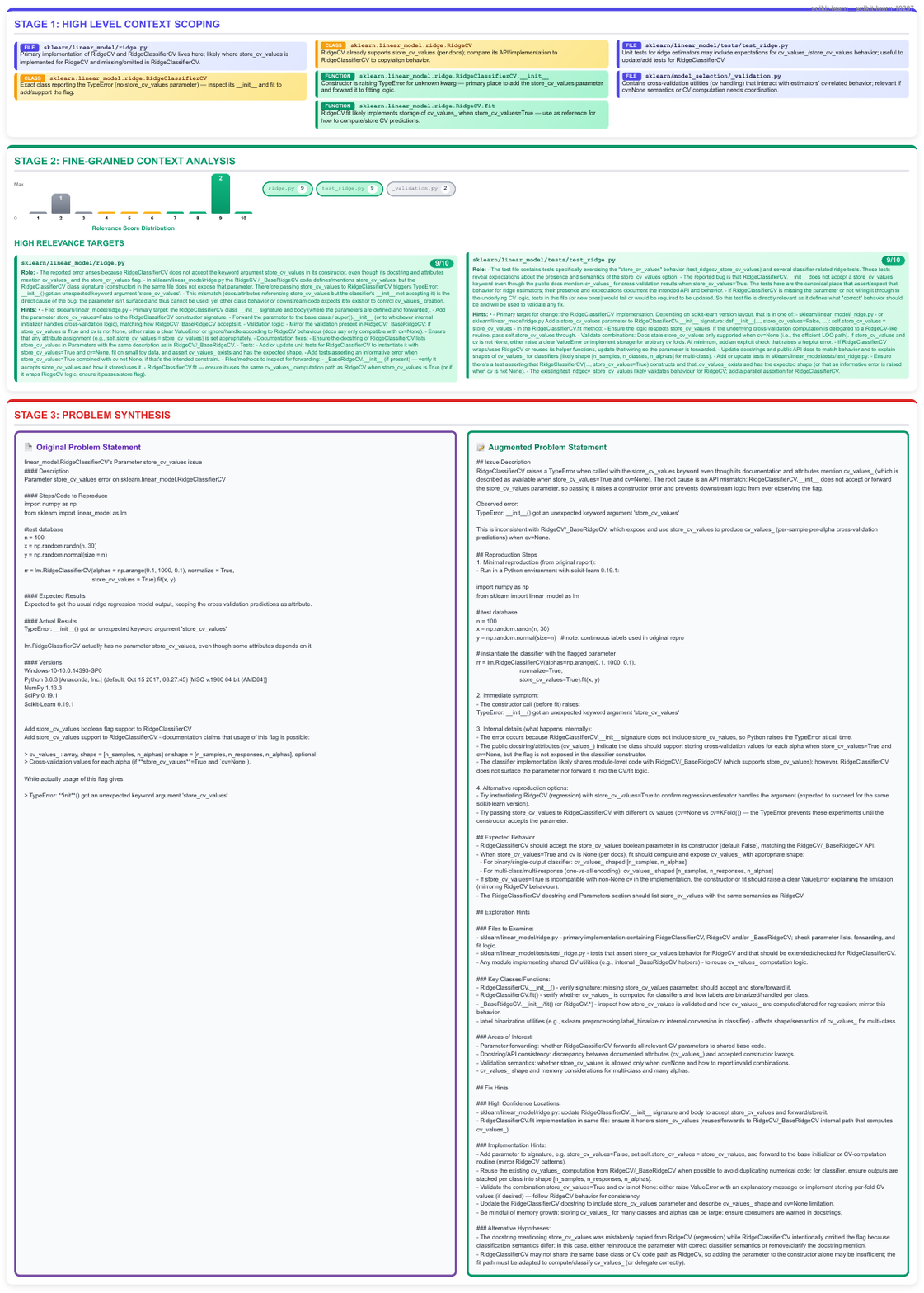}
\caption{CodeScout visualization for \texttt{scikit-learn\_\_scikit-learn-10297} with GPT-5-mini.}
\label{fig:sklearn-gpt5}
\end{figure*}

\begin{figure*}
\centering
\includegraphics[width=\textwidth,page=1]{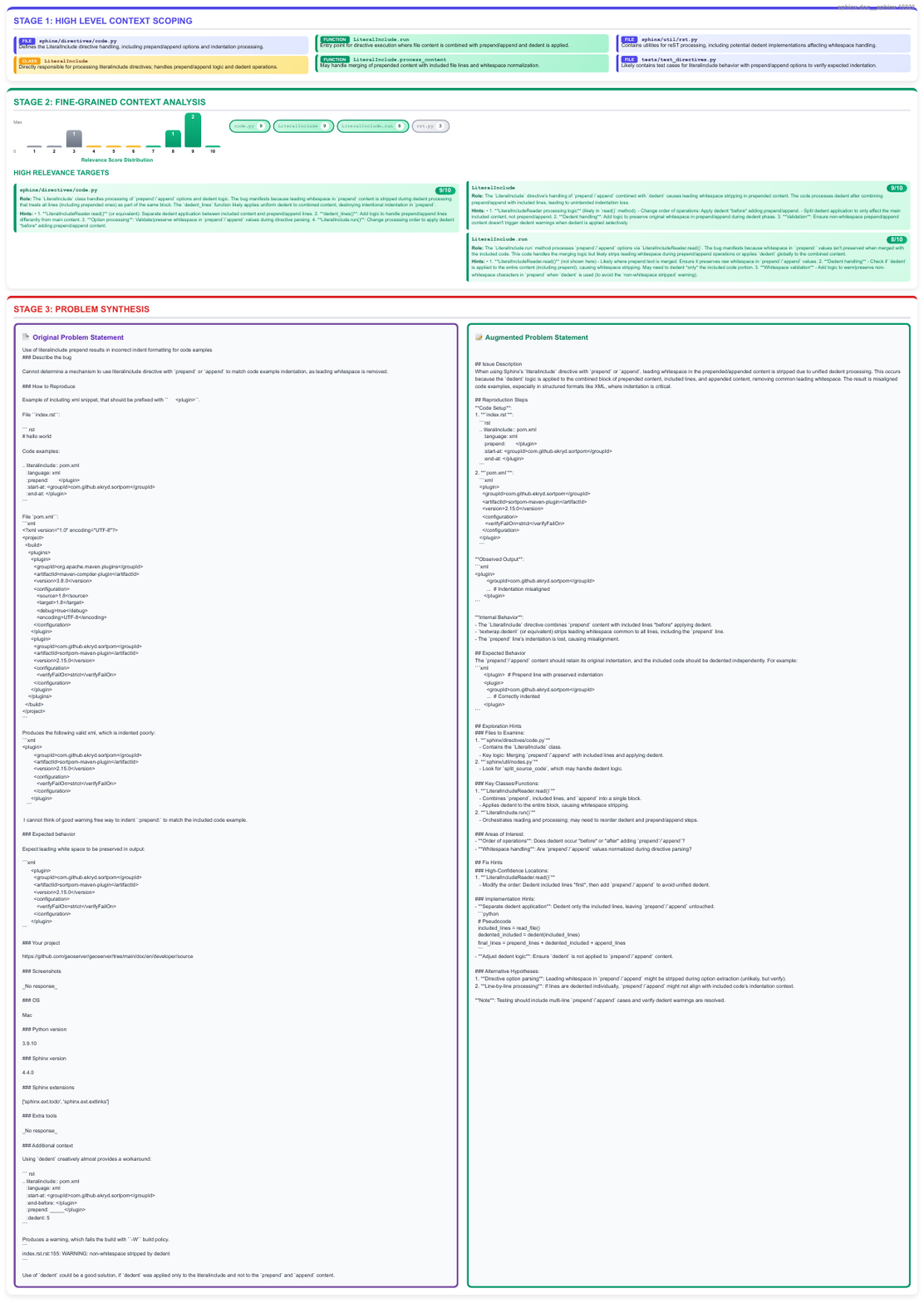}
\caption{CodeScout visualization for \texttt{sphinx-doc\_\_sphinx-10323} with DeepSeek-R1.}
\label{fig:sphinx-dsr1}
\end{figure*}

\begin{figure*}
\centering
\includegraphics[width=\textwidth,page=1]{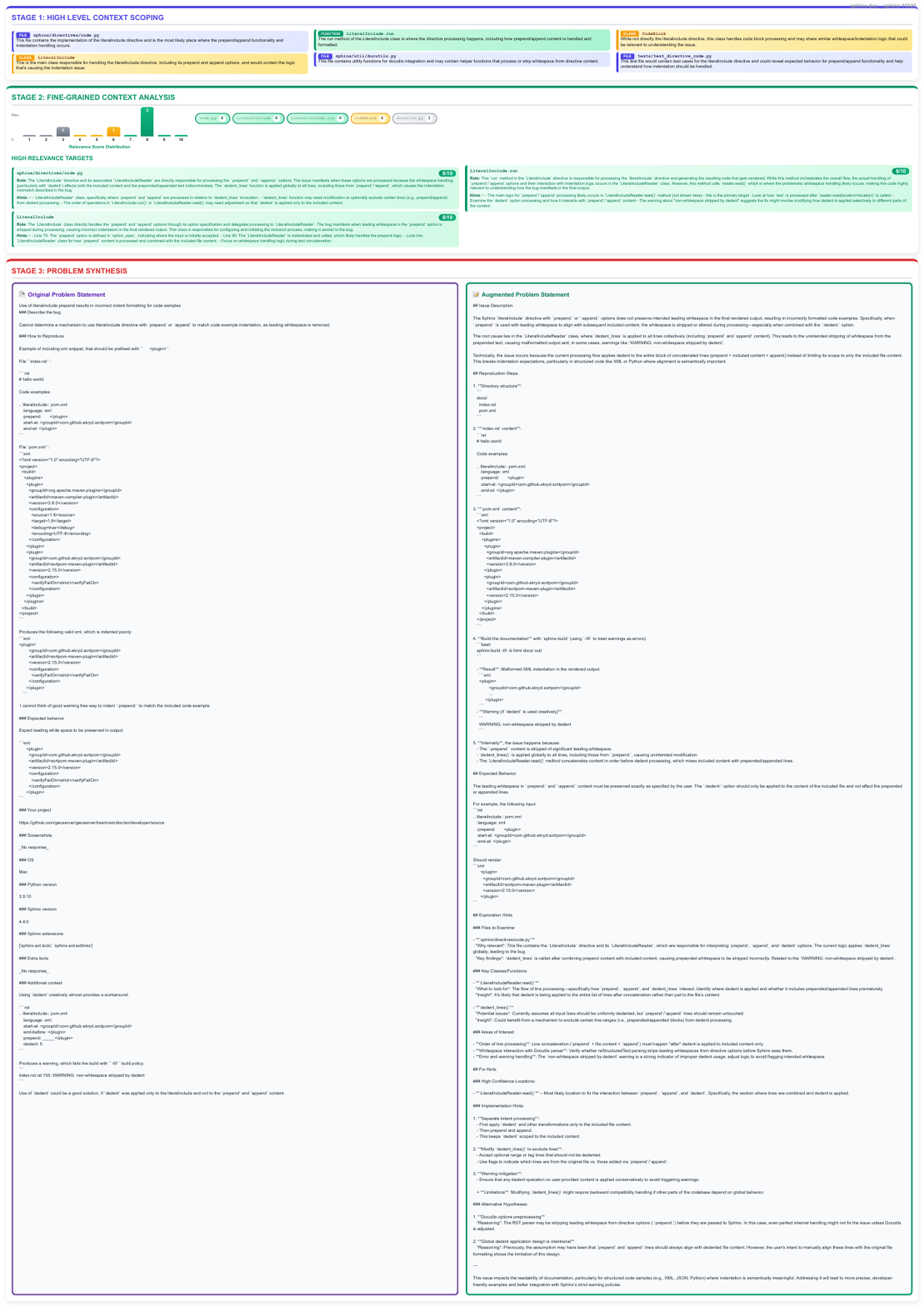}
\caption{CodeScout visualization for \texttt{sphinx-doc\_\_sphinx-10323} with Qwen3-Coder.}
\label{fig:sphinx-q3c}
\end{figure*}

\begin{figure*}
\centering
\includegraphics[width=\textwidth,page=1]{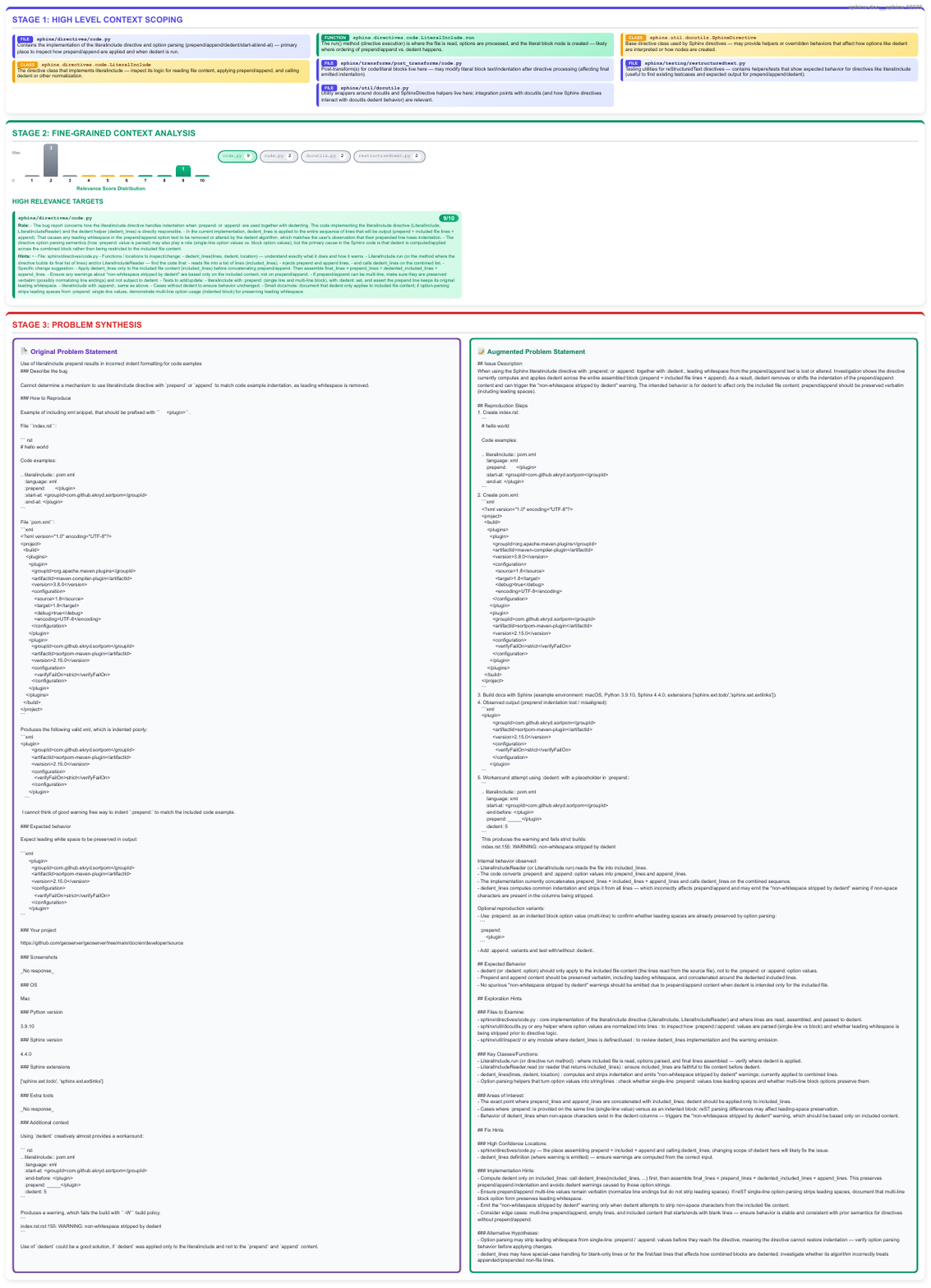}
\caption{CodeScout visualization for \texttt{sphinx-doc\_\_sphinx-10323} with GPT-5-mini.}
\label{fig:sphinx-gpt5}
\end{figure*}

\begin{figure*}
\centering
\includegraphics[width=\textwidth,page=1]{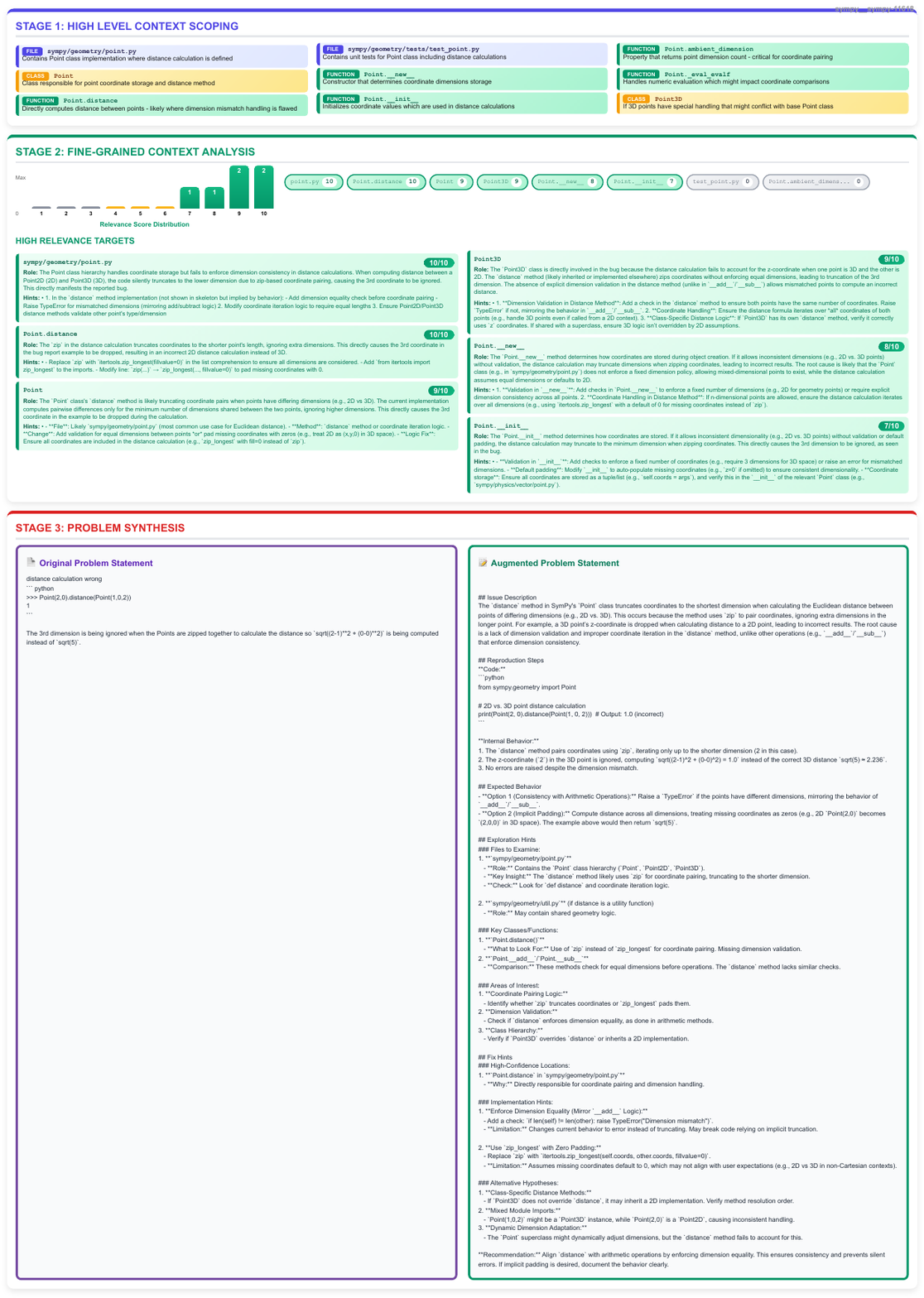}
\caption{CodeScout visualization for \texttt{sympy\_\_sympy-11618} with DeepSeek-R1.}
\label{fig:sympy-dsr1}
\end{figure*}

\begin{figure*}
\centering
\includegraphics[width=\textwidth,page=1]{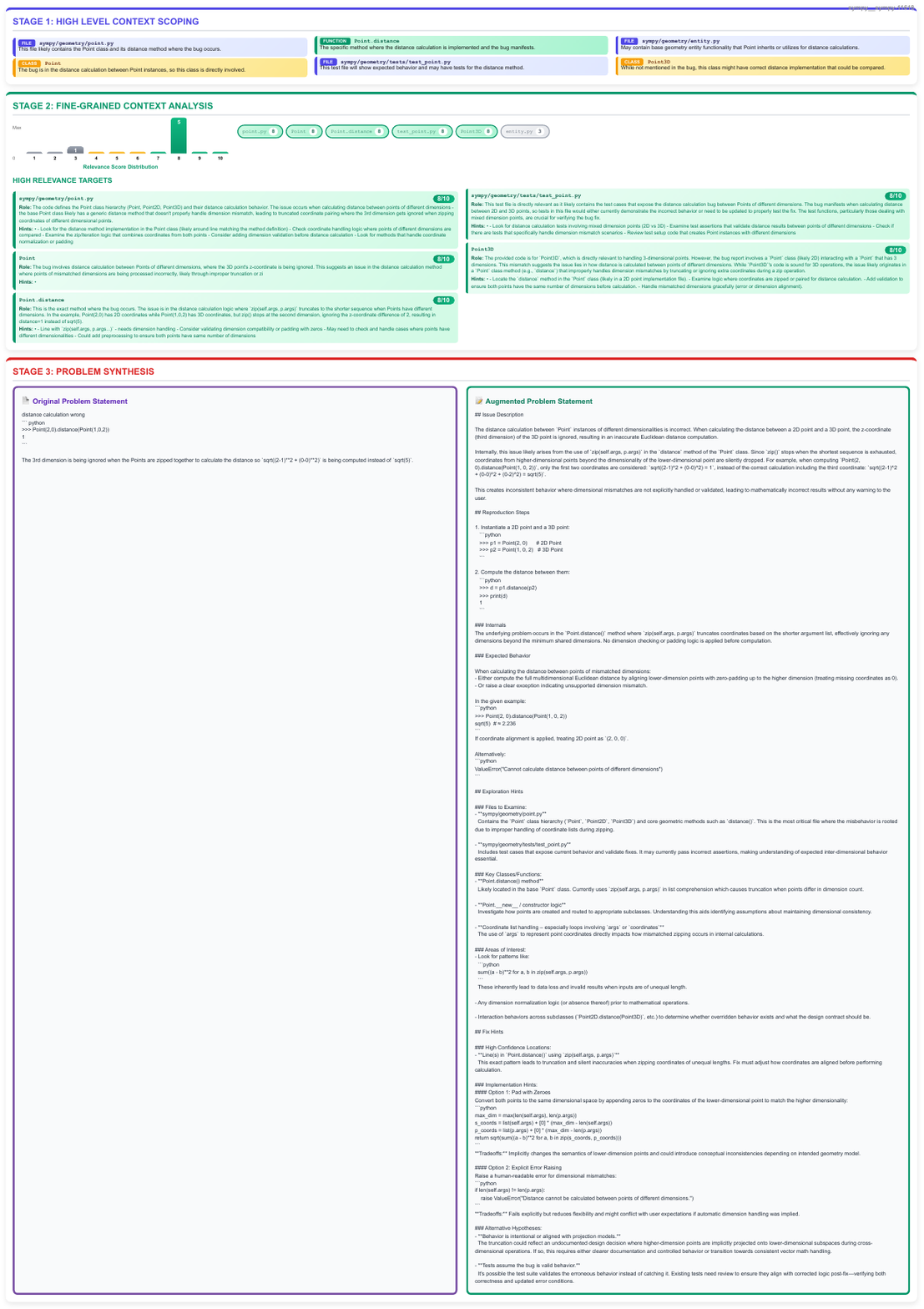}
\caption{CodeScout visualization for \texttt{sympy\_\_sympy-11618} with Qwen3-Coder.}
\label{fig:sympy-q3c}
\end{figure*}

\begin{figure*}
\centering
\includegraphics[width=\textwidth,page=1]{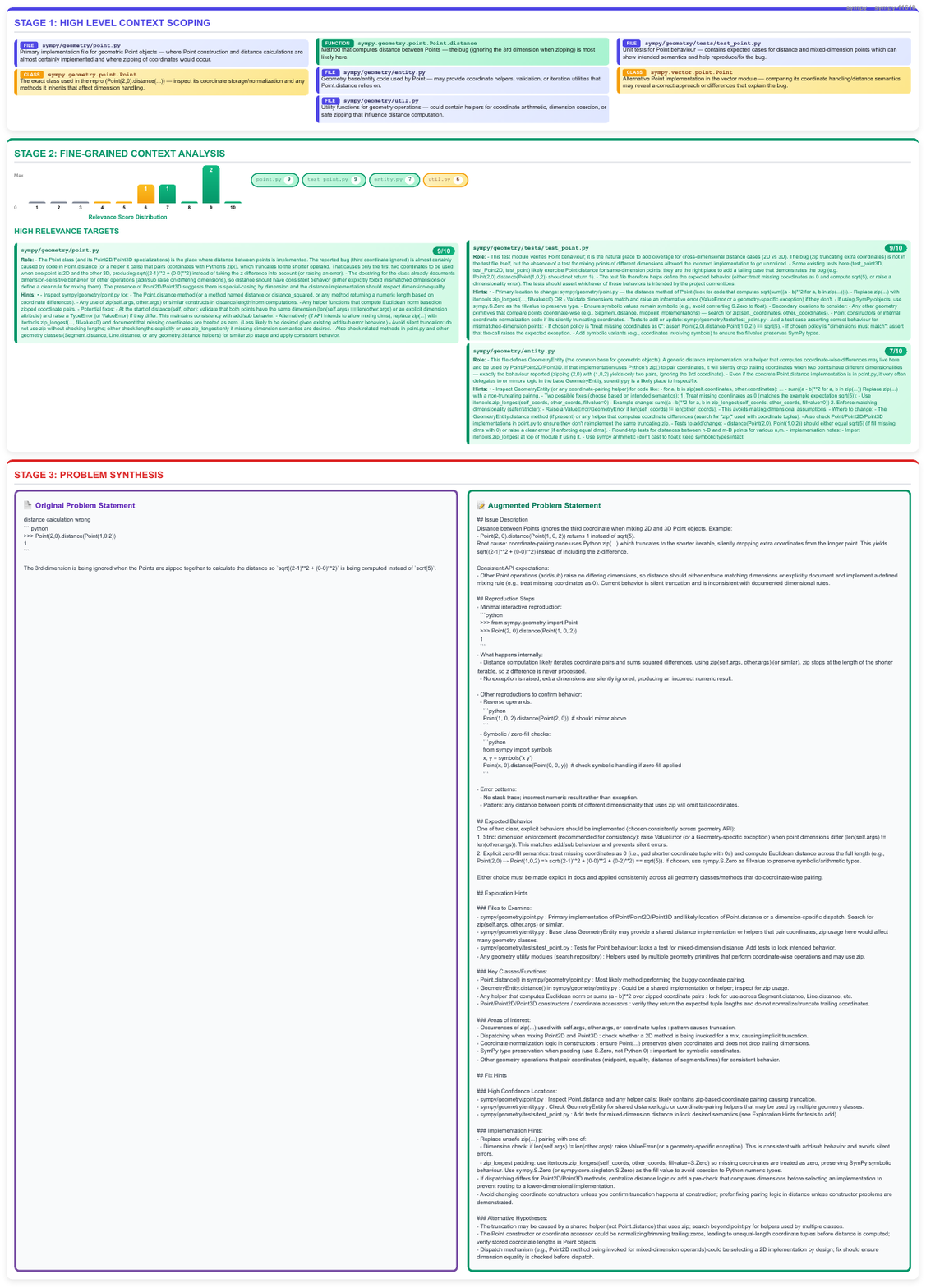}
\caption{CodeScout visualization for \texttt{sympy\_\_sympy-11618} with GPT-5-mini.}
\label{fig:sympy-gpt5}
\end{figure*}

\subsection{Quantitaive Comparison}
We present comprehensive comparative analysis of the CodeScout pipeline for the chosen LLMs: DeepSeek-R1, Qwen3-Coder, and GPT-5-mini.

\begin{figure}
\centering
\includegraphics[width=\columnwidth]{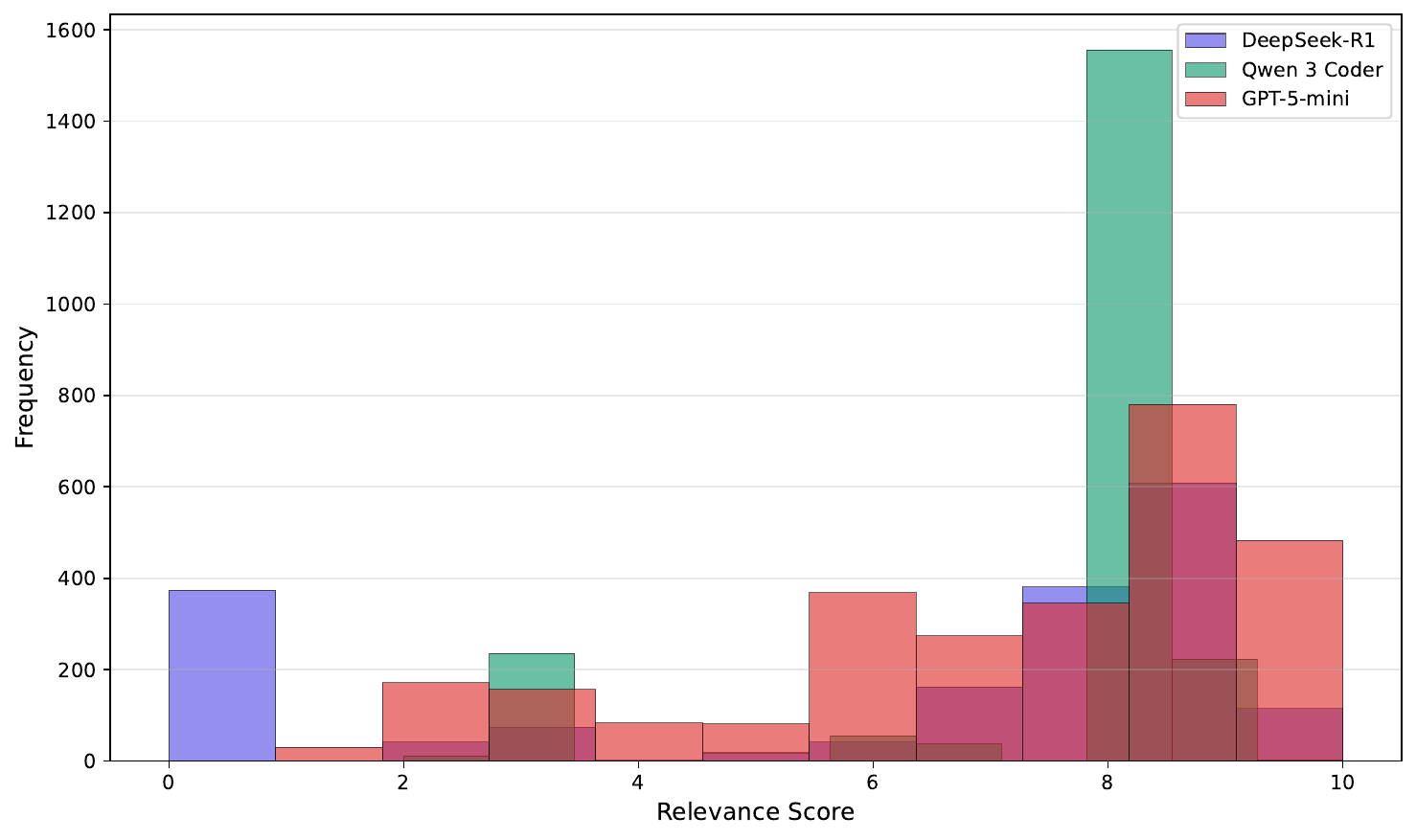}
\caption{CodeScout relevance score distribution across all three methods.}
\label{fig:score-distribution}
\end{figure}

\begin{figure}
\centering
\includegraphics[width=\columnwidth]{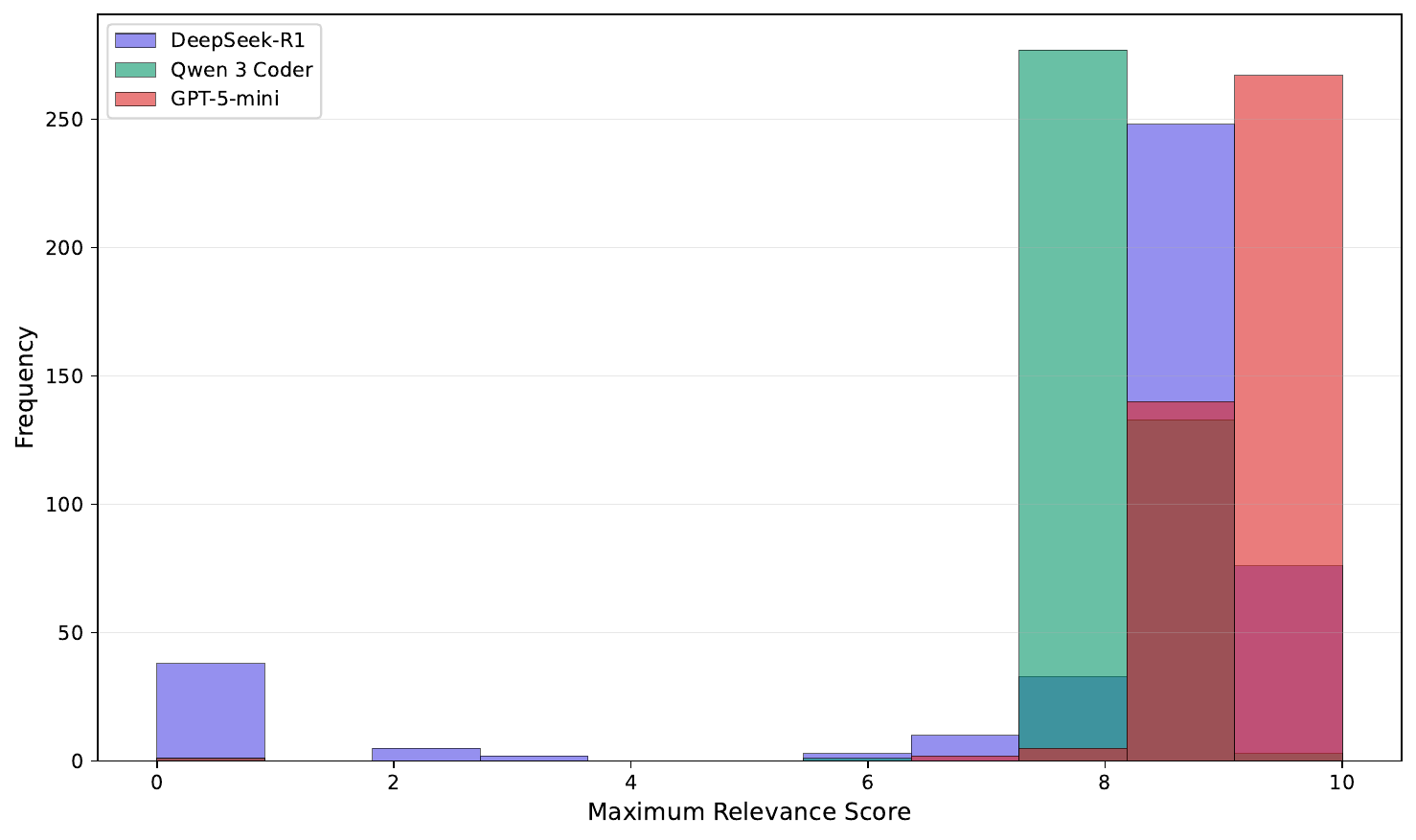}
\caption{CodeScout maximum relevance score distribution per instance.}
\label{fig:max-score-distribution}
\end{figure}

\textbf{Score Distributions.} Figure~\ref{fig:score-distribution} shows the distribution of relevance scores assigned by each method across all targets. Qwen3-Coder and GPT-5-mini show higher concentrations of scores in the 7-9 range, while DeepSeek-R1 exhibits a broader distribution with more scores in the lower ranges. Figure~\ref{fig:max-score-distribution} displays the distribution of maximum scores per instance, showing that Qwen3-Coder and GPT-5-mini more frequently assign maximum scores of 8-9, whereas DeepSeek-R1 shows greater variability.

\begin{figure}
\centering
\includegraphics[width=\columnwidth]{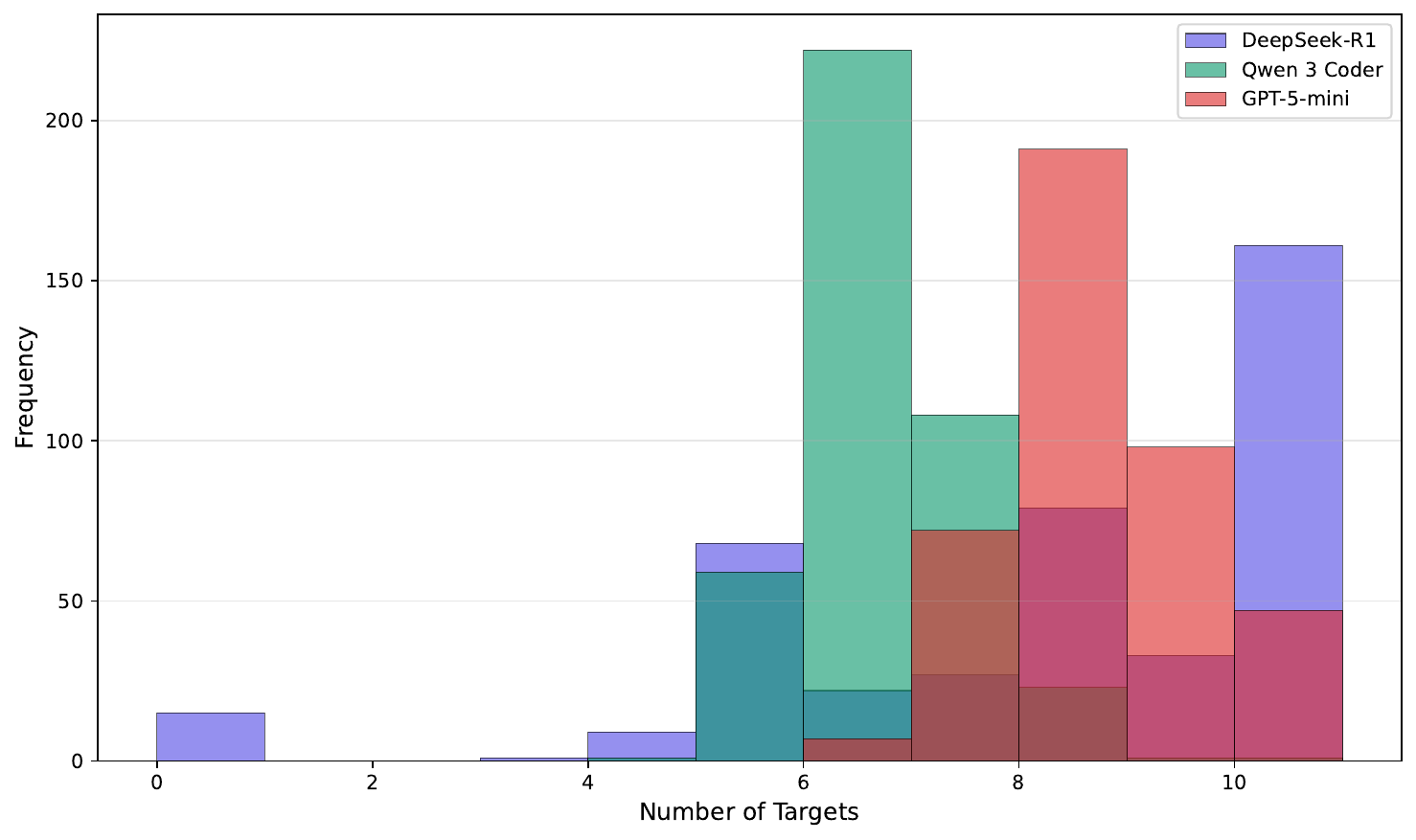}
\caption{CodeScout number of targets distribution per instance.}
\label{fig:num-targets}
\end{figure}

\begin{figure}
\centering
\includegraphics[width=\columnwidth]{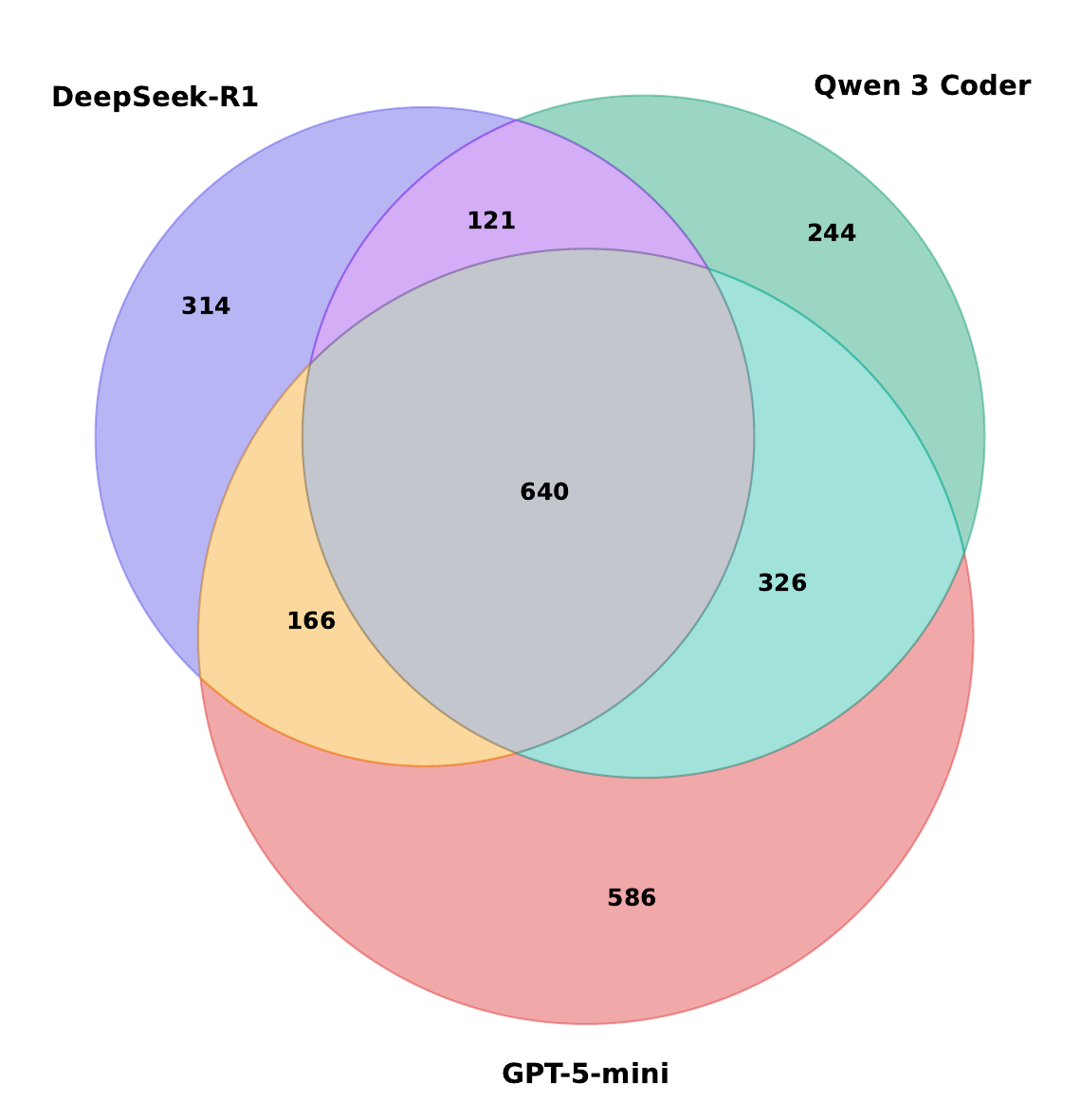}
\caption{CodeScout target coverage: Venn diagram showing unique and shared exploration targets.}
\label{fig:venn-targets}
\end{figure}

\textbf{Target Coverage.} Figure~\ref{fig:num-targets} presents the distribution of the number of targets identified per instance for each method. DeepSeek-R1 tends to identify more targets per instance with a broader distribution, while Qwen3-Coder shows a tighter distribution around 6 targets. Figure~\ref{fig:venn-targets} illustrates the overlap of unique targets identified across methods, showing that a substantial portion of targets are method-specific, with GPT-5-mini identifying the largest unique set.

\begin{figure*}
\centering
\begin{subfigure}[b]{0.32\textwidth}
\centering
\includegraphics[width=\textwidth]{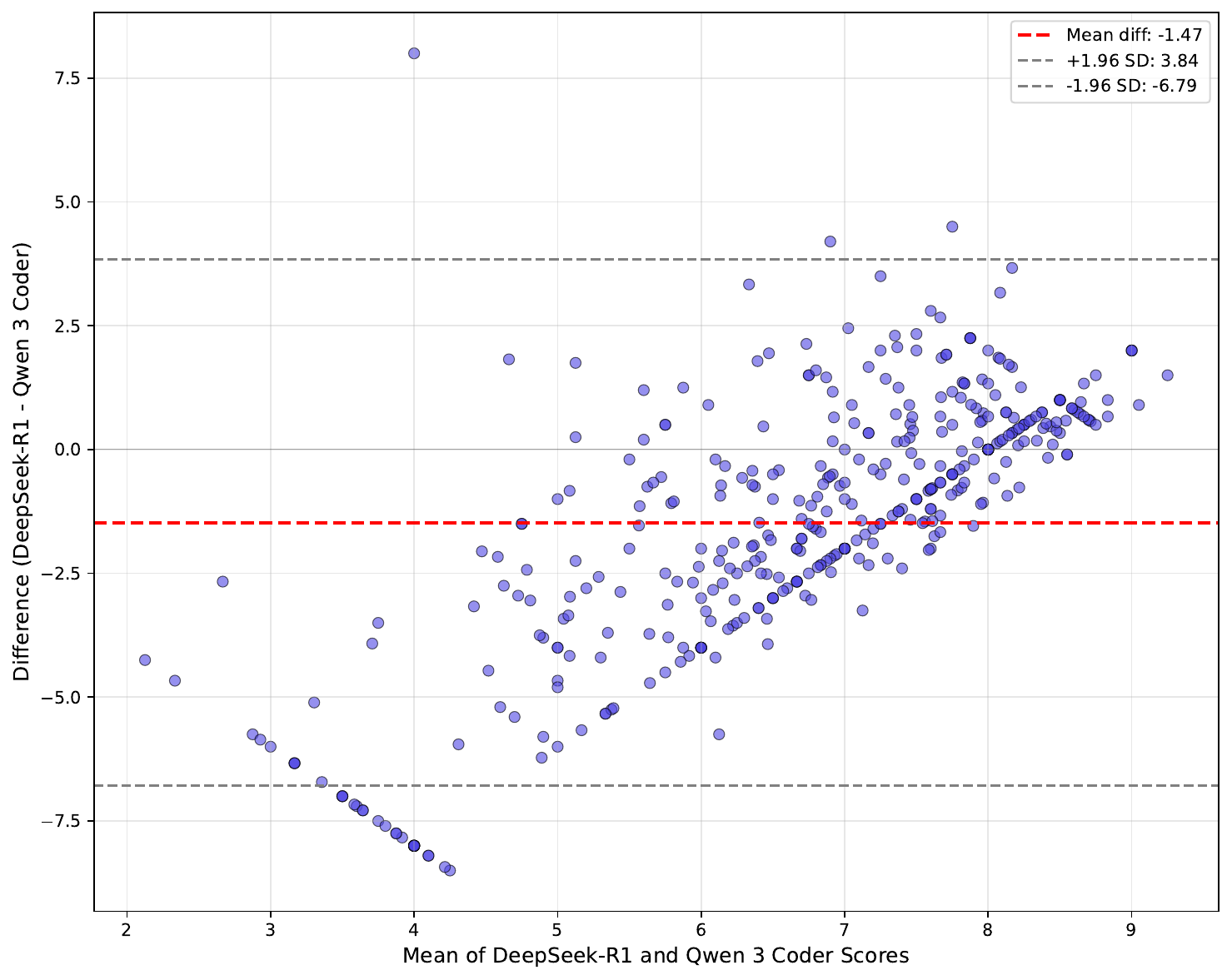}
\caption{DeepSeek-R1 vs Qwen3-Coder}
\label{fig:bland-altman-dsr1-q3c}
\end{subfigure}
\hfill
\begin{subfigure}[b]{0.32\textwidth}
\centering
\includegraphics[width=\textwidth]{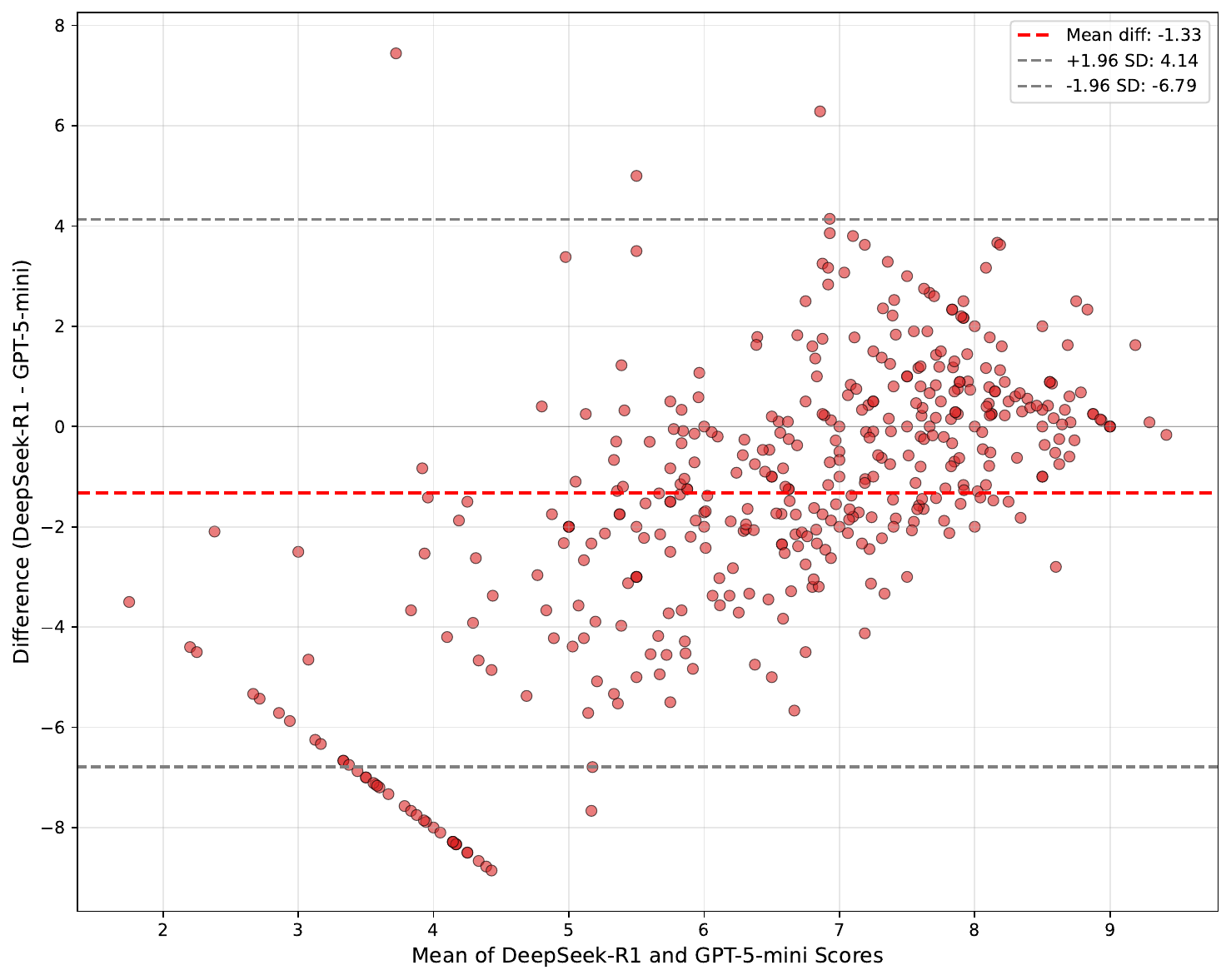}
\caption{DeepSeek-R1 vs GPT-5-mini}
\label{fig:bland-altman-dsr1-gpt5}
\end{subfigure}
\hfill
\begin{subfigure}[b]{0.32\textwidth}
\centering
\includegraphics[width=\textwidth]{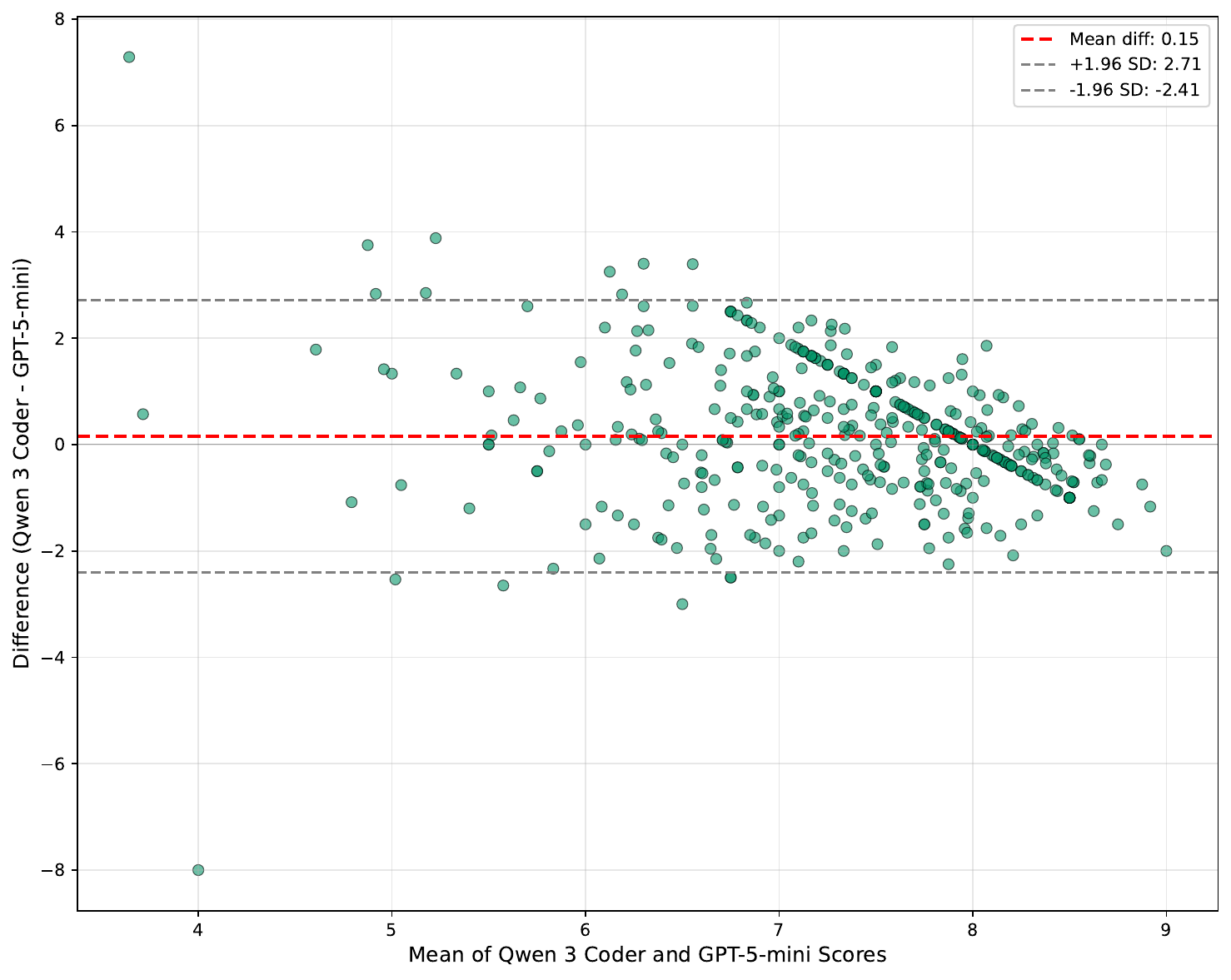}
\caption{Qwen3-Coder vs GPT-5-mini}
\label{fig:bland-altman-q3c-gpt5}
\end{subfigure}
\caption{CodeScout agreement analysis: Bland-Altman plots comparing average scores between LLMs.}
\label{fig:bland-altman-all}
\end{figure*}

\textbf{Agreement Analysis.} Figure~\ref{fig:bland-altman-all} presents Bland-Altman plots comparing the average scores between method pairs. These plots reveal systematic differences: DeepSeek-R1 consistently scores lower than both Qwen3-Coder and GPT-5-mini, as evidenced by the negative mean differences. The agreement limits show the range of score differences across instances.

\begin{figure*}
\centering
\begin{subfigure}[b]{0.32\textwidth}
\centering
\includegraphics[width=\textwidth]{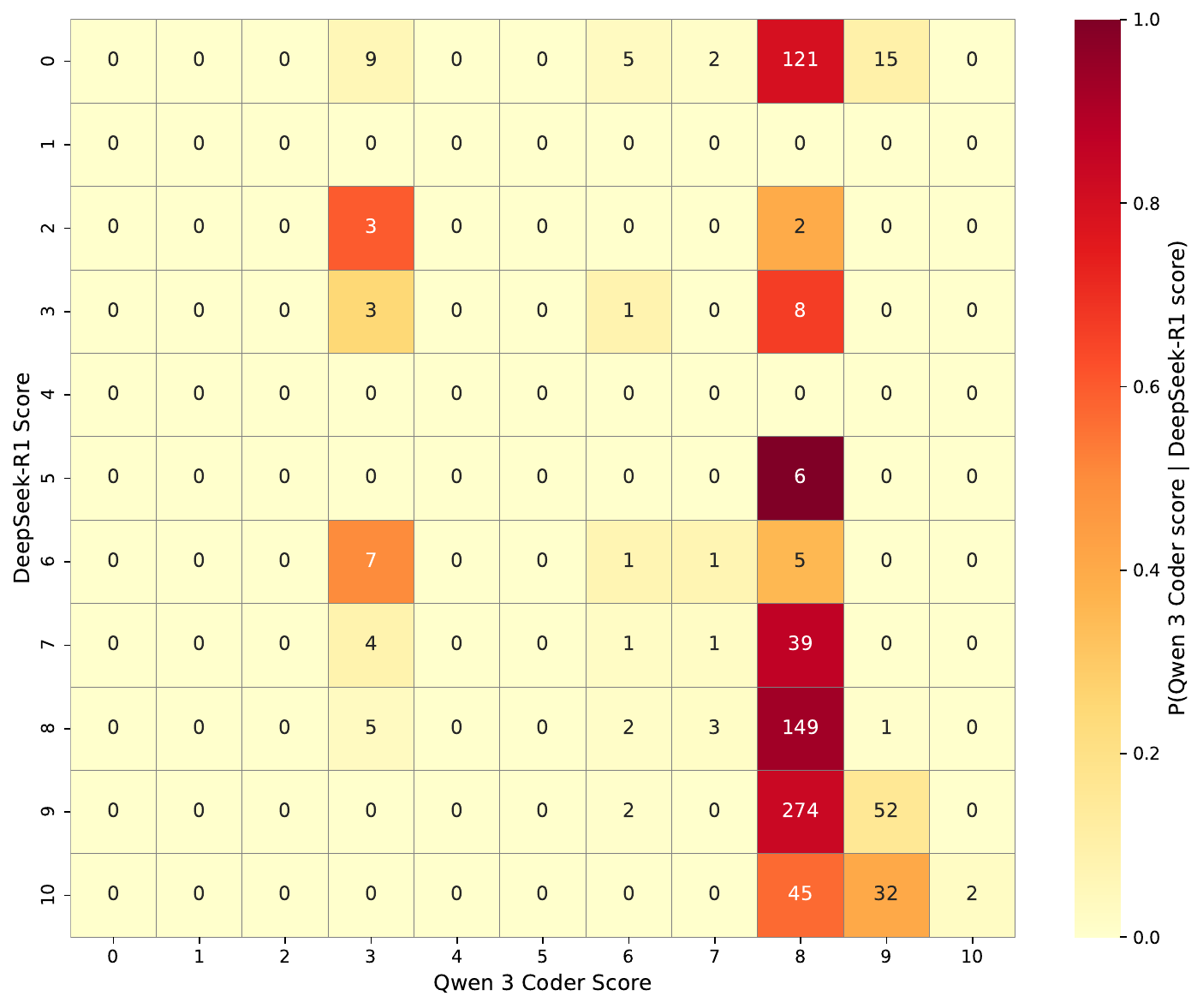}
\caption{DeepSeek-R1 vs Qwen3-Coder}
\label{fig:agreement-dsr1-q3c}
\end{subfigure}
\hfill
\begin{subfigure}[b]{0.32\textwidth}
\centering
\includegraphics[width=\textwidth]{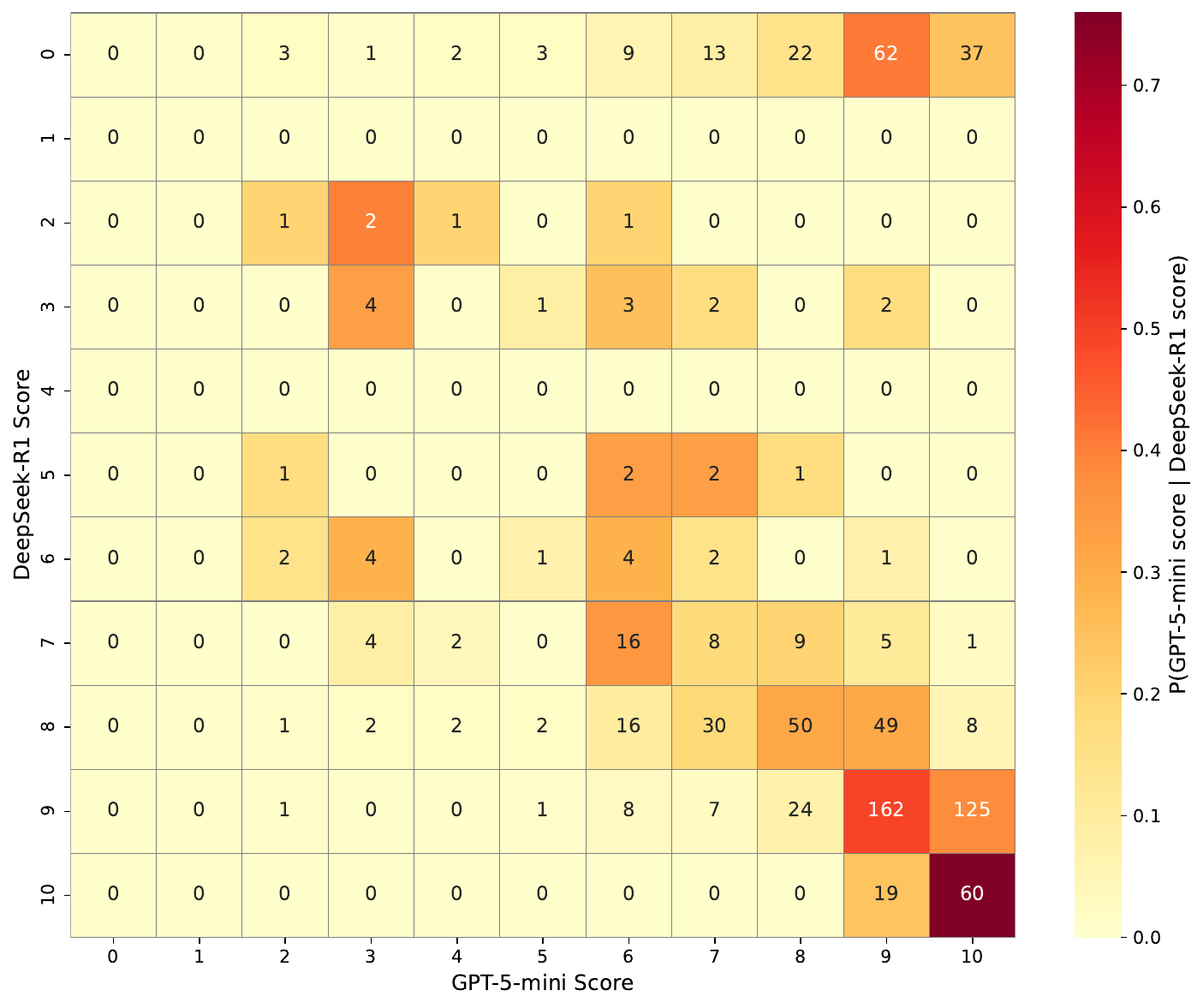}
\caption{DeepSeek-R1 vs GPT-5-mini}
\label{fig:agreement-dsr1-gpt5}
\end{subfigure}
\hfill
\begin{subfigure}[b]{0.32\textwidth}
\centering
\includegraphics[width=\textwidth]{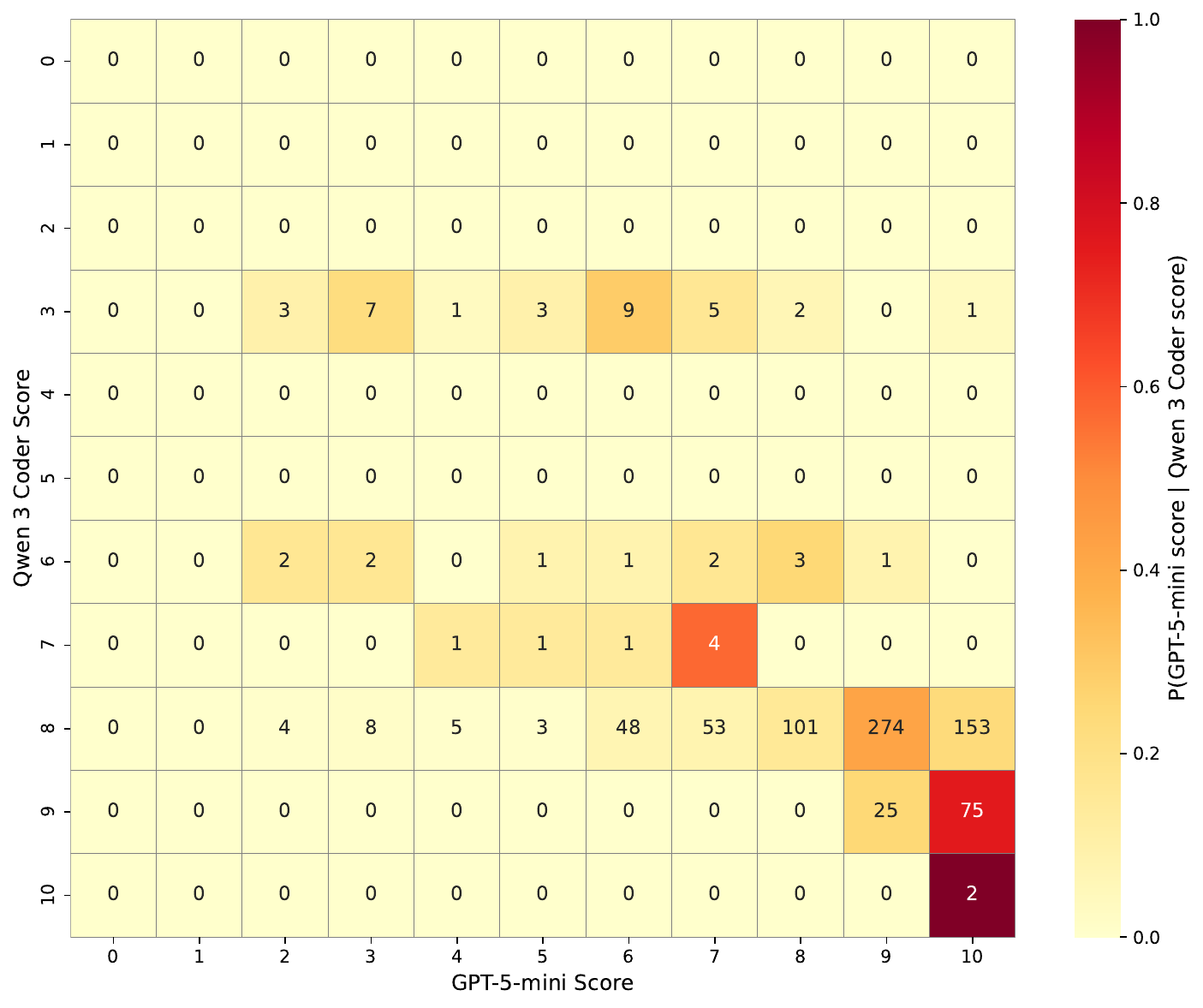}
\caption{Qwen3-Coder vs GPT-5-mini}
\label{fig:agreement-q3c-gpt5}
\end{subfigure}
\caption{CodeScout score agreement heatmaps between methods for common targets.}
\label{fig:agreement-all}
\end{figure*}

\textbf{Score Correlations.} Figure~\ref{fig:agreement-all} shows heatmaps of score agreement for targets that were analyzed by multiple methods. Each cell shows the count of targets where the row method assigned one score and the column method assigned another. The diagonal entries represent exact agreement, while off-diagonal entries indicate disagreement. The conditional probabilities in the colorbar show how likely one method is to assign a particular score given the other method's score.

\begin{thebibliography}{24}
\providecommand{\natexlab}[1]{#1}

\bibitem[{Ahmed et~al.(2024)Ahmed, Pai, Devanbu, and Barr}]{ahmed2024automatic}
Toufique Ahmed, Kunal~Suresh Pai, Premkumar Devanbu, and Earl~T Barr. 2024.
\newblock Automatic semantic augmentation of language model prompts (for code summarization).
\newblock pages 1--13.

\bibitem[{Barke et~al.(2023)Barke, James, and Polikarpova}]{barke2023grounded}
Shraddha Barke, Michael~B. James, and Nadia Polikarpova. 2023.
\newblock \href {https://doi.org/10.1145/3586030} {Grounded copilot: How programmers interact with code-generating models}.
\newblock \emph{Proceedings of the ACM on Programming Languages}, 7(OOPSLA1):85--111.

\bibitem[{Bouzenia et~al.(2025)Bouzenia, Devanbu, and Pradel}]{bouzenia2025repairagent}
Islem Bouzenia, Premkumar Devanbu, and Michael Pradel. 2025.
\newblock Repairagent: An autonomous, llm-based agent for program repair.
\newblock In \emph{Proceedings of the 47th International Conference on Software Engineering}, ICSE '25, pages 2188--2200. IEEE Computer Society.

\bibitem[{Bouzenia and Pradel(2025)}]{bouzenia2025understanding}
Islem Bouzenia and Michael Pradel. 2025.
\newblock Understanding software engineering agents: A study of thought-action-result trajectories.
\newblock \emph{arXiv preprint arXiv:2506.18824}.

\bibitem[{Jiang et~al.(2023)Jiang, Liu, Lutellier, and Tan}]{jiang2023impact}
Nan Jiang, Kevin Liu, Thibaud Lutellier, and Lin Tan. 2023.
\newblock \href {https://doi.org/10.1109/ICSE48619.2023.00125} {Impact of code language models on automated program repair}.
\newblock In \emph{Proceedings of the 45th International Conference on Software Engineering}, ICSE '23, pages 1430--1442. IEEE Press.

\bibitem[{Jimenez et~al.(2023)Jimenez, Yang, Wettig, Yao, Pei, Press, and Narasimhan}]{jimenez2023swe}
Carlos~E. Jimenez, John Yang, Alexander Wettig, Shunyu Yao, Kexin Pei, Ofir Press, and Karthik Narasimhan. 2023.
\newblock \href {https://arxiv.org/abs/2310.06770} {Swe-bench: Can language models resolve real-world github issues?}
\newblock \emph{Preprint}, arXiv:2310.06770.

\bibitem[{Jin et~al.(2023)Jin, Shahriar, Tufano, Shi, Lu, Sundaresan, and Svyatkovskiy}]{jin2023inferfixendtoendprogramrepair}
Matthew Jin, Syed Shahriar, Michele Tufano, Xin Shi, Shuai Lu, Neel Sundaresan, and Alexey Svyatkovskiy. 2023.
\newblock \href {https://arxiv.org/abs/2303.07263} {Inferfix: End-to-end program repair with llms}.
\newblock \emph{Preprint}, arXiv:2303.07263.

\bibitem[{Kumar et~al.(2025)Kumar, Khare, Sharma, Kumar, Saini, Yadav, Jain, Rana, Verma, Meena, and Edubilli}]{kumar2025intuitionevidencemeasuringais}
Anand Kumar, Vishal Khare, Deepak Sharma, Satyam Kumar, Vijay Saini, Anshul Yadav, Sachendra Jain, Ankit Rana, Pratham Verma, Vaibhav Meena, and Avinash Edubilli. 2025.
\newblock \href {https://arxiv.org/abs/2509.19708} {Intuition to evidence: Measuring ai's true impact on developer productivity}.
\newblock \emph{Preprint}, arXiv:2509.19708.

\bibitem[{Lemieux et~al.(2023)Lemieux, Inala, Lahiri, and Sen}]{lemieux2023codamosa}
Caroline Lemieux, Jeevana~Priya Inala, Shuvendu~K. Lahiri, and Siddhartha Sen. 2023.
\newblock \href {https://doi.org/10.1109/ICSE48619.2023.00085} {Codamosa: Escaping coverage plateaus in test generation with pre-trained large language models}.
\newblock In \emph{Proceedings of the 45th International Conference on Software Engineering}, pages 919--931, Melbourne, Australia. IEEE/ACM.

\bibitem[{Liu et~al.(2024)Liu, Wang, Chen, Peng, Chen, Zhang, and Lou}]{liu2024large}
Junwei Liu, Kaixin Wang, Yixuan Chen, Xin Peng, Zhenpeng Chen, Lingming Zhang, and Yiling Lou. 2024.
\newblock \href {https://arxiv.org/abs/2409.02977} {Large language model-based agents for software engineering: A survey}.
\newblock \emph{Preprint}, arXiv:2409.02977.

\bibitem[{Lu et~al.(2015)Lu, Sun, Wang, Lo, and Duan}]{lu2015query}
Meili Lu, Xiaobing Sun, Shaowei Wang, David Lo, and Yucong Duan. 2015.
\newblock Query expansion via wordnet for effective code search.
\newblock \emph{2015 IEEE 22nd International Conference on Software Analysis, Evolution, and Reengineering (SANER)}, pages 545--549.

\bibitem[{Meng et~al.(2024)Meng, Ma, Gao, and Peng}]{meng2024empirical}
Xiangxin Meng, Zexiong Ma, Pengfei Gao, and Chao Peng. 2024.
\newblock An empirical study on llm-based agents for automated bug fixing.
\newblock \emph{arXiv preprint arXiv:2411.10213}.

\bibitem[{Nam et~al.(2025)Nam, Omran, Murillo, Thakur, Araujo, Blistein, Fr{\"o}mmgen, Hellendoorn, and Chandra}]{nam2025prompting}
Daye Nam, Ahmed Omran, Ambar Murillo, Saksham Thakur, Abner Araujo, Marcel Blistein, Alexander Fr{\"o}mmgen, Vincent Hellendoorn, and Satish Chandra. 2025.
\newblock Prompting llms for code editing: Struggles and remedies.
\newblock \emph{arXiv preprint arXiv:2504.20196}.

\bibitem[{Nie et~al.(2016)Nie, Jiang, Ren, Sun, and Li}]{nie2016query}
Liming Nie, He~Jiang, Zhilei Ren, Zeyi Sun, and Xiaochen Li. 2016.
\newblock Query expansion based on crowd knowledge for code search.
\newblock \emph{IEEE Transactions on Services Computing}, 9:771--783.

\bibitem[{Robertson et~al.(2009)Robertson, Zaragoza et~al.}]{Bm25}
Stephen Robertson, Hugo Zaragoza, and 1 others. 2009.
\newblock The probabilistic relevance framework: Bm25 and beyond.
\newblock \emph{Foundations and Trends{\textregistered} in Information Retrieval}, 3(4):333--389.

\bibitem[{Romeo et~al.(2025)Romeo, Arena, Blefari, Pironti, Lupinacci, and Furfaro}]{romeo2025arpaccino}
Francesco Romeo, Luigi Arena, Francesco Blefari, Francesco~Aurelio Pironti, Matteo Lupinacci, and Angelo Furfaro. 2025.
\newblock \href {https://arxiv.org/abs/2507.10584} {Arpaccino: An agentic-rag for policy as code compliance}.
\newblock \emph{Preprint}, arXiv:2507.10584.

\bibitem[{Ryan et~al.(2024)Ryan, Jain, Shang, Wang, Ma, Ramanathan, and Ray}]{ryan2024code}
Gabriel Ryan, Siddhartha Jain, Mingyue Shang, Shiqi Wang, Xiaofei Ma, Murali~Krishna Ramanathan, and Baishakhi Ray. 2024.
\newblock \href {https://doi.org/10.1145/3643769} {Code-aware prompting: A study of coverage-guided test generation in regression setting using llm}.
\newblock \emph{Proceedings of the ACM on Software Engineering}, 1(FSE):951--971.

\bibitem[{Wang et~al.(2025)Wang, Li, Song, Xu, Tang, Zhuge, Pan, Song, Li, Singh, Tran, Li, Ma, Zheng, Qian, Shao, Muennighoff, Zhang, Hui, Lin, Brennan, Peng, Ji, and Neubig}]{wang2025openhands}
Xingyao Wang, Boxuan Li, Yufan Song, Frank~F. Xu, Xiangru Tang, Mingchen Zhuge, Jiayi Pan, Yueqi Song, Bowen Li, Jaskirat Singh, Hoang~H. Tran, Fuqiang Li, Ren Ma, Mingzhang Zheng, Bill Qian, Yanjun Shao, Niklas Muennighoff, Yizhe Zhang, Binyuan Hui, and 5 others. 2025.
\newblock \href {https://openreview.net/forum?id=OJd3ayDDoF} {Openhands: An open platform for {AI} software developers as generalist agents}.
\newblock In \emph{The Thirteenth International Conference on Learning Representations}.

\bibitem[{Wang et~al.(2023)Wang, Guo, Shi, Chen, Chen, Zhong, Wang, Li, Zhang, Lyu, and Zheng}]{wang2023you}
Yanlin Wang, Lianghong Guo, Ensheng Shi, Wenqing Chen, Jiachi Chen, Wanjun Zhong, Menghan Wang, Hui Li, Hongyu Zhang, Ziyu Lyu, and Zibin Zheng. 2023.
\newblock You augment me: Exploring chatgpt-based data augmentation for semantic code search.
\newblock \emph{2023 IEEE International Conference on Software Maintenance and Evolution (ICSME)}, pages 14--25.

\bibitem[{Wang and Xu(2024)}]{wang2024srsa}
Yaqi Wang and Haipei Xu. 2024.
\newblock \href {https://arxiv.org/abs/2411.14574} {Srsa: A cost-efficient strategy-router search agent for real-world human-machine interactions}.
\newblock \emph{Preprint}, arXiv:2411.14574.

\bibitem[{Yang et~al.(2024)Yang, Jimenez, Wettig, Lieret, Yao, Narasimhan, and Press}]{yang2024sweagent}
John Yang, Carlos~E. Jimenez, Alexander Wettig, Kilian Lieret, Shunyu Yao, Karthik Narasimhan, and Ofir Press. 2024.
\newblock {SWE}-agent: Agent-computer interfaces enable automated software engineering.
\newblock In \emph{Advances in Neural Information Processing Systems}, volume~37.

\bibitem[{Yuan et~al.(2024)Yuan, Liu, Ding, Wang, Chen, Peng, and Lou}]{yuan2024evaluating}
Zhiqiang Yuan, Mingwei Liu, Shiji Ding, Kaixin Wang, Yixuan Chen, Xin Peng, and Yiling Lou. 2024.
\newblock \href {https://doi.org/10.1145/3660783} {Evaluating and improving {ChatGPT} for unit test generation}.
\newblock \emph{Proceedings of the ACM on Software Engineering}, 1(FSE):1703--1726.

\bibitem[{Zhang et~al.(2024)Zhang, Ruan, Fan, and Roychoudhury}]{zhang2024autocoderover}
Yuntong Zhang, Haifeng Ruan, Zhiyu Fan, and Abhik Roychoudhury. 2024.
\newblock \href {https://doi.org/10.1145/3650212.3680384} {Autocoderover: Autonomous program improvement}.
\newblock In \emph{Proceedings of the 33rd ACM SIGSOFT International Symposium on Software Testing and Analysis}, ISSTA 2024, pages 1592--1604, New York, NY, USA. Association for Computing Machinery.

\bibitem[{Ziegler et~al.(2022)Ziegler, Kalliamvakou, Simister, Sittampalam, Li, Rice, Rifkin, and Aftandilian}]{ziegler2022productivity}
Albert Ziegler, Eirini Kalliamvakou, Shawn Simister, Ganesh Sittampalam, X.~Alice Li, Andrew Rice, Devon Rifkin, and Edward Aftandilian. 2022.
\newblock \href {https://doi.org/10.1145/3520312.3534864} {Productivity assessment of neural code completion}.
\newblock In \emph{Proceedings of the 6th ACM SIGPLAN International Symposium on Machine Programming}, MAPS 2022, New York, NY, USA. Association for Computing Machinery.

\end{thebibliography}
\end{document}